\documentclass[pdflatex,iicol,sn-apa]{sn-jnl}
\usepackage{graphicx}%
\usepackage{multirow}%
\usepackage{amsmath,amssymb,amsfonts}%
\usepackage{amsthm}%
\usepackage[title]{appendix}%
\usepackage{xcolor}%
\usepackage{textcomp}%
\usepackage{manyfoot}%
\usepackage{booktabs}%
\usepackage{algorithm}%
\usepackage{algorithmicx}%
\usepackage{algpseudocode}%
\usepackage{listings}%
\theoremstyle{thmstyleone}%
\theoremstyle{thmstyletwo}%
\theoremstyle{thmstylethree}%
\usepackage{array}
\usepackage{colortbl}
\usepackage{pifont}
\usepackage{tabularx}

\definecolor{EgoAIMRow}{RGB}{238,238,238}
\definecolor{HOILeft}{RGB}{80,150,215}
\definecolor{HOIRight}{RGB}{245,140,150}

\providecommand{\tblYes}{\ding{51}}
\providecommand{\tblNo}{\ding{55}}

\providecommand{\tblMark}[1]{%
  \rlap{\textsuperscript{\ensuremath{#1}}}%
}

\providecommand{\tblHead}[1]{{\small #1}}

\newsavebox{\tblMeasureBox}

\usepackage{etoolbox}

\let\OriginalTitlefont\Titlefont
\renewcommand{\Titlefont}{%
  \OriginalTitlefont
  \fontsize{20bp}{30bp}\selectfont
}

\let\OriginalAbstractfont\abstractfont
\renewcommand{\abstractfont}{%
  \OriginalAbstractfont
  \linespread{1.2}\selectfont
}

\makeatletter

\patchcmd{\@maketitle}
  {\else\vskip21pt\fi}
  {\else\vskip7pt\fi}
  {}
  {\PackageWarning{title-spacing}{Title top-space patch failed}}

\patchcmd{\@maketitle}
  {\artauthors\par}
  {\vspace{6pt}\artauthors\par\vspace{16pt}}
  {}
  {\PackageWarning{title-spacing}{Title-block spacing patch failed}}

\makeatother

\patchcmd{\keywords}
  {\addvspace{10pt}}
  {\addvspace{16pt}}
  {}
  {\PackageWarning{keyword-spacing}{Keywords spacing patch failed}}

\makeatletter
\AtBeginDocument{\NAT@longnamesfalse}
\makeatother

\AtBeginDocument{%
  \apptocmd{\thebibliography}{%
    \let\OriginalBibNewblock\newblock
    \renewcommand{\newblock}{%
      \ifhmode
        \OriginalBibNewblock
      \fi
    }%
  }%
  {}%
  {\PackageWarning{bib-spacing}{Bibliography patch failed}}%
}

\begin{document}
\title[Gaze2HOI]{Gaze2HOI: Gaze-Aware Intention Disambiguation for Egocentric Hand-Object Interaction Forecasting}

\author[1]{\fnm{Chuanyang} \sur{Ma}}\nomail
\author[2]{\fnm{Chenye} \sur{Wang}}\nomail
\author[1]{\fnm{Yixuan} \sur{Wang}}\nomail
\author[1]{\fnm{Huiling} \sur{Zhou}}\nomail
\author[1]{\fnm{Man} \sur{Zhang}}\nomail
\author[3]{\fnm{Xingqun} \sur{Qi}}\nomail

\author*[1]{\fnm{Muyi} \sur{Sun}}
\email{muyi.sun@bupt.edu.cn}

\author[4]{\fnm{Caifeng} \sur{Shan}}\nomail
\author[5]{\fnm{Zhenan} \sur{Sun}}\nomail

\affil[1]{%
    \,\orgname{Beijing University of Posts and Telecommunications},
    \orgaddress{\city{Beijing}, \country{China}}}

\affil[2]{%
    \,\orgname{The Hong Kong University of Science and Technology},
    \orgaddress{\city{Hong Kong}, \country{China}}}

\affil[3]{%
    \,\orgname{North China Electric Power University},
    \orgaddress{\city{Beijing}, \country{China}}}

\affil[4]{%
    \,\orgdiv{Suzhou Campus},
    \orgname{Nanjing University},
    \orgaddress{\city{Suzhou}, \country{China}}}

\affil[5]{%
    \,\orgdiv{Institute of Automation},
    \orgname{Chinese Academy of Sciences},
    \orgaddress{\city{Beijing}, \country{China}}}

\abstract{
Anticipating how users will interact with their surroundings is essential for proactive systems that provide timely assistance in extended reality and embodied applications.
Prior work typically forecasts hand-object interactions directly from historical observations, leaving intention and contact geometry implicit.
However, similar observation histories can precede semantically or geometrically distinct futures.
Inspired by behavioral evidence that gaze can provide early cues for underlying intention, we propose Gaze2HOI, a gaze-aware framework that formulates hand-object interaction forecasting as a hierarchical problem comprising intention inference, contact grounding, and coupled hand-object motion generation.
Specifically, Gaze2HOI first weights gaze cues by fixation strength and recency, and uses them to modulate the visual representations of a vision-language model (VLM) through Temporal Gaze Saliency Modulation (TGSM), thereby reducing semantic ambiguity in intention inference.
Subsequently, an Anticipatory Contact Grounding (ACG) module maps these structured intentions into explicit contact, support, and placement fields, effectively mitigating geometric ambiguity.
Driven by these explicit conditions, a diffusion-based model jointly generates the six-degree-of-freedom (6-DoF) object trajectories and future articulated motion of both hands.
To evaluate Gaze2HOI, we further introduce EgoAIM, a phase-aware benchmark that organizes egocentric hand-object interactions into four phases defined by evolving hand-object relations and provides fine-grained intention annotations.
Extensive experiments demonstrate that Gaze2HOI achieves state-of-the-art performance in both intention inference and motion forecasting, even approaching oracle-level end-to-end performance in specific interaction phases.
Project page: \url{https://gaze2hoi.github.io/}.
}


\maketitle

\section{Introduction}
\begin{figure}[!t]
    \centering
    \includegraphics[width=\columnwidth]{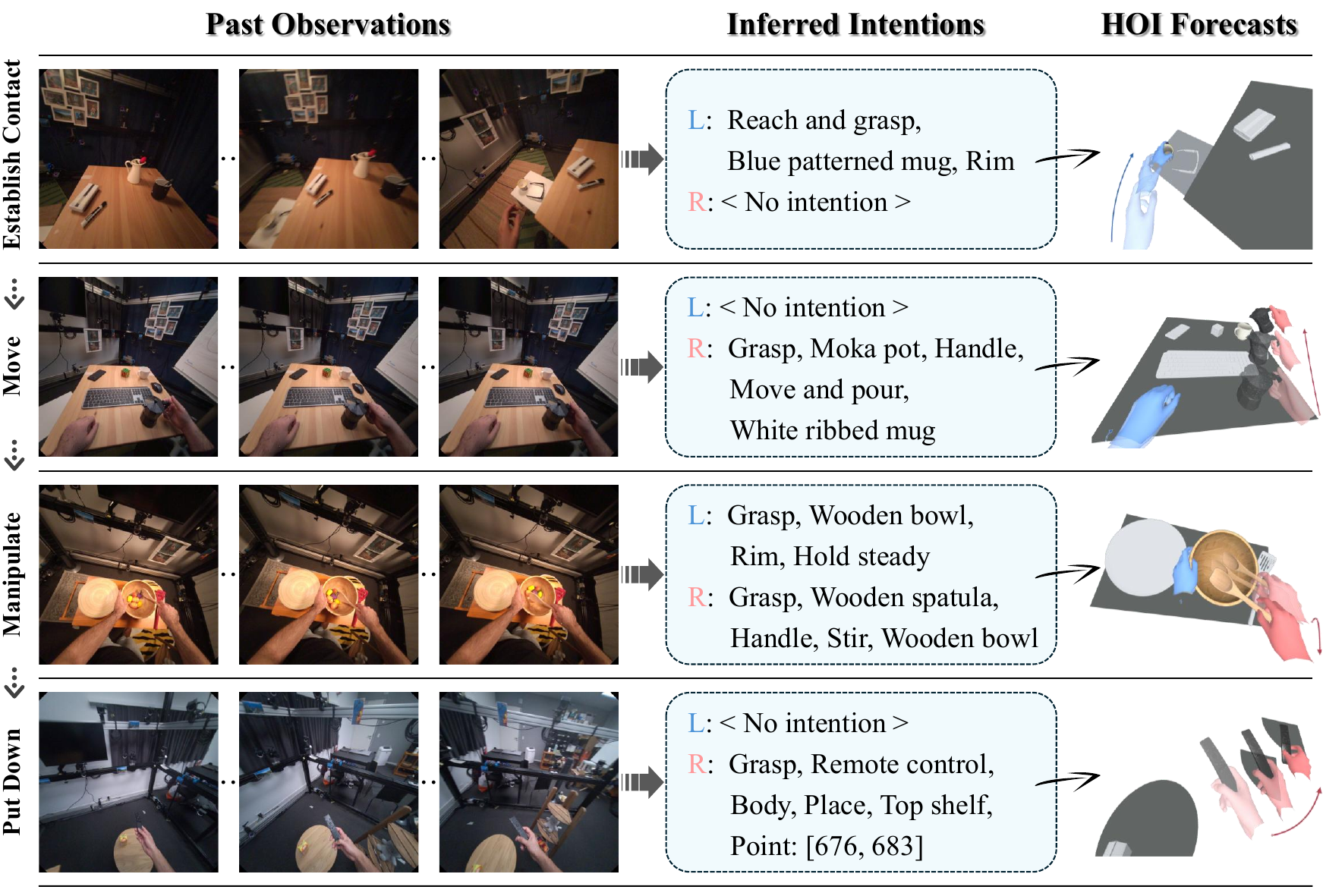}
    \caption{\textbf{\emph{Egocentric hand-object interaction forecasting.}}
    Gaze2HOI forecasts coordinated interactions involving both hands and multiple objects across four phases: \emph{Establish Contact}, \emph{Move}, \emph{Manipulate}, and \emph{Put Down}.
    \textcolor{HOILeft}{L} and \textcolor{HOIRight}{R} denote the left and right hands, respectively.}
    \label{fig1}
\end{figure}

Recent advances in extended reality and embodied intelligence have enabled interactive systems to tackle increasingly complex tasks, spurring a growing demand for natural and anticipatory assistance~\citep{Driess2023PaLM-E,Zheng2022GIMO}.
However, many existing systems still operate under a reactive paradigm, typically responding to explicit commands or actions that are already underway~\citep{Lee2025SensibleAgent}.
To achieve seamless human-machine collaboration, systems must \textbf{transition from passive responders to proactive assistants} capable of anticipating human intentions and upcoming actions~\citep{Li2022EgoPAT3D}.
Since routine manipulation is largely realized through hand interactions with surrounding objects, egocentric hand-object interaction (HOI) forecasting has become a foundational task that bridges perceptual input and proactive human-machine assistance~\citep{Plizzari2024EgocentricSurvey,Liu2022OCT,Bao2023USST,Liu2024TACO}.
Figure~\ref{fig1} illustrates this task through representative forecasts across interaction phases.

Existing methods have explored various aspects of egocentric HOI forecasting but still struggle to capture continuous, coordinated hand-object interaction dynamics~\citep{Liu2024TACO}.
Early approaches primarily anticipate discrete action targets or events, identifying interaction types or locations without recovering the underlying continuous kinematics~\citep{Xu2023DCR,Furnari2019RULSTM,Li2022EgoPAT3D,Pasca2024TransFusion}.
To capture spatial evolution, subsequent research extends these predictions to continuous trajectories by forecasting either future hand positions~\citep{Liu2022OCT,Bao2023USST,Hatano2025EgoH4} or isolated object poses~\citep{Soraki2026ObjectForesight}.
While these methods model the temporal evolution of individual elements, they fail to jointly model fine-grained finger articulation and object motion, significantly limiting their capacity to represent complex hand-object coordination~\citep{Ma2025Diff-IP2D}.
Recent efforts in joint hand-object forecasting attempt to bridge this coordination gap. 
However, these approaches typically learn a direct mapping from historical observations to future trajectories, leaving human intention and physical contact geometry implicit~\citep{Liu2024TACO,Corona2020ContextAware}.
Since similar historical observations can precede completely divergent futures~\citep{Stergiou2025AboutTime,Liu2022OCT}, such direct-mapping paradigms struggle to determine a coherent interaction when faced with multiple plausible continuations~\citep{Liu2024TACO}.

\begin{figure*}[!t]
    \centering
    \includegraphics[width=\textwidth]{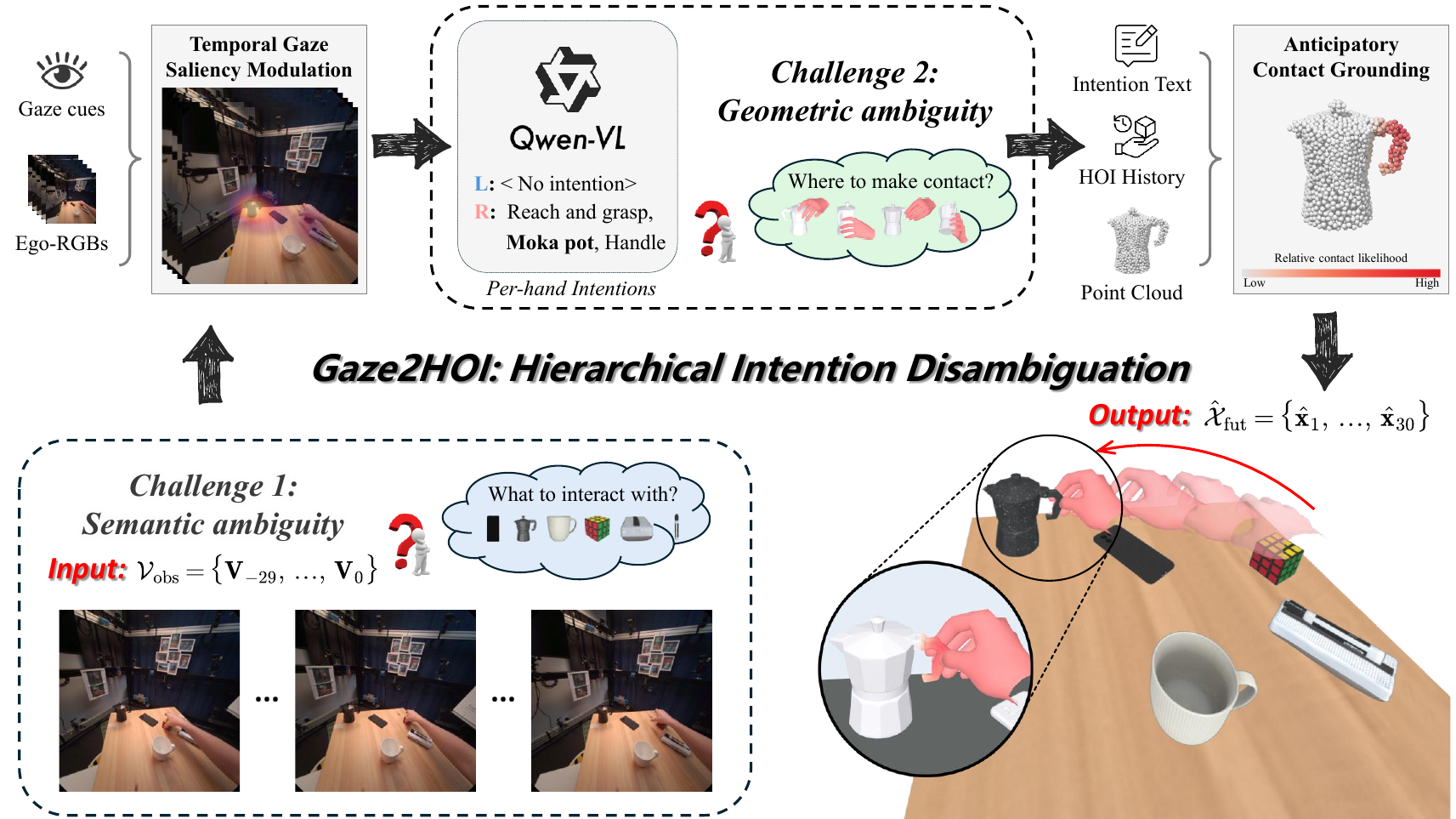}
    \caption{\textbf{\emph{Hierarchical intention disambiguation in Gaze2HOI.}} Gaze-aware intention inference reduces semantic ambiguity about the intended object and action. Anticipatory Contact Grounding then combines the inferred intentions with object geometry and interaction history to localize likely contact regions, reducing geometric ambiguity. The resulting semantic and geometric cues guide coordinated hand-object motion forecasting.}
    \label{fig2}
    \vspace{0.5 mm}
\end{figure*}

Reliable forecasting of coordinated hand-object motion requires the resolution of two intertwined ambiguities: \textbf{semantic ambiguity} in the intended interaction and \textbf{geometric ambiguity} in its physical realization.
As illustrated in Figure~\ref{fig2}, similar observation histories can be compatible with different intended actions, while even the same action involving the same object can exhibit multiple valid contact configurations.

To address these challenges, we propose Gaze2HOI, a gaze-aware framework that formulates hand-object interaction forecasting as a hierarchical problem comprising \textbf{intention inference, contact grounding, and coupled hand-object motion generation}.
Inspired by behavioral evidence that gaze can provide early cues before an intention fully manifests in hand motion~\citep{Johansson2001EyeHandCoordination}, we first introduce Temporal Gaze Saliency Modulation (TGSM).
This module weights gaze cues according to fixation strength and recency to modulate the visual representation of a VLM (e.g., Qwen3-VL~\citep{Bai2025Qwen3VL}), thereby reducing semantic ambiguity by explicitly identifying the intended object and action.
Meanwhile, an inferred intention such as “reach and grasp the moka pot handle” still leaves the precise contact configuration unspecified~\citep{Christen2024DiffH2O}.
To resolve this geometric ambiguity, we employ an Anticipatory Contact Grounding (ACG) module that combines the inferred intentions with object geometry and historical interaction information to predict contact, support, and placement fields.
These fields express the anticipated interaction through spatially explicit relationships, providing geometric guidance for a conditional diffusion model that jointly forecasts the 6-DoF object trajectories and future articulated motion of both hands. 

This proposed hierarchical formulation calls for a benchmark that pairs fine-grained future intention annotations with continuous hand-object motion.
As hand-object relations evolve, the cues available for intention inference and the demands of motion coordination also change~\citep{Johansson2001EyeHandCoordination,He2025GazeGuided}.
Therefore, \textbf{phase-specific evaluation can reveal performance differences obscured by aggregate metrics}.
Existing benchmarks for action anticipation primarily assess future action categories or interaction targets~\citep{Damen2022EPICKITCHENS100,Grauman2022Ego4D,Li2022EgoPAT3D}, whereas those for 3D hand-object tracking and motion forecasting emphasize spatial states and their temporal evolution~\citep{Banerjee2025HOT3D,Liu2024TACO}.
However, these task-specific protocols provide limited support for jointly assessing intention inference and coordinated motion forecasting under these changing interaction conditions.
To support these evaluations, we introduce \textbf{EgoAIM}, a phase-aware benchmark for \underline{Ego}centric \underline{A}nticipation of \underline{I}ntention and \underline{M}otion.
Built on existing egocentric recordings~\citep{Banerjee2025HOT3D}, EgoAIM comprises 7,429 forecasting windows organized into four phases defined by evolving hand-object relations.
EgoAIM pairs the fine-grained future intention annotations for each hand with continuous 3D hand-object motion and provides furniture support-plane annotations to characterize the environmental geometry relevant to object support and placement.
This design supports evaluating intention inference and motion forecasting both separately and jointly across the four interaction phases.

Our contributions are summarized as follows:
\begin{itemize}
    \item We reformulate the task of egocentric hand-object interaction forecasting as a \textbf{hierarchical problem} comprising intention inference, contact grounding, and coupled hand-object motion generation.

    \item We introduce \textbf{EgoAIM}, a phase-aware benchmark with 7,429 forecasting windows, featuring fine-grained intention and geometry-grounded annotations for egocentric hand-object interaction forecasting.

    \item We propose \textbf{Gaze2HOI}, coupling Temporal Gaze Saliency Modulation and Anticipatory Contact Grounding for semantic and geometric disambiguation to drive a diffusion-based model for hand-object forecasting.

    \item We extensively evaluate Gaze2HOI in both stage-wise and overall system settings, demonstrating state-of-the-art performance in intention inference and motion forecasting across interaction phases.
\end{itemize}

\section{Related Work}
This section reviews egocentric hand-object interaction forecasting as the task context. 
It also examines two complementary directions: conditional motion generation and gaze-aware intention understanding for inferring semantic conditions from historical observations.

\subsection{Egocentric Hand-Object Interaction Forecasting}
Research on egocentric hand-object interaction forecasting spans various levels of prediction granularity.
Target-oriented approaches anticipate action locations~\citep{Li2022EgoPAT3D} or next-active objects alongside associated actions and time-to-contact~\citep{Pasca2024TransFusion,Dessalene2023EgoOMG}, providing spatial or semantic summaries without recovering articulated hand-object dynamics.
To capture temporal evolution, probabilistic approaches jointly predict 2D hand trajectories and interaction hotspots~\citep{Liu2020FHOI,Liu2022OCT}, while Diff-IP2D and MADiff incorporate camera egomotion into diffusion-based trajectory prediction~\citep{Ma2025Diff-IP2D,Ma2026MADiff}.
Extending to three-dimensional space, USST models uncertainty in future hand positions~\citep{Bao2023USST}, whereas Uni-Hand predicts hand-center and selected-joint trajectories together with contact and separation states~\citep{Ma2026Uni-Hand}.
Focusing on fine-grained articulation, HandDiff forecasts single-hand joint motion through diffusion~\citep{Tong2026HandDiff}, while \citet{He2025GazeGuided} combine historical hand motion, gaze cues, and object locations to forecast the articulated motion of both hands in egocentric grasping tasks.
ObjectForesight focuses on object dynamics, using scene geometry and historical poses to forecast future 6-DoF trajectories~\citep{Soraki2026ObjectForesight}.
Although these methods capture increasingly detailed motion, their predictions typically isolate hands from objects, failing to jointly model their coordinated evolution.
Moving toward joint forecasting, TACO benchmarks the simultaneous prediction of articulated bimanual motion and the poses of tools and target objects~\citep{Liu2024TACO}.
Despite these advances, existing methods generally map historical observations directly to future motion, leaving interaction intention and 3D contact geometry implicit~\citep{Liu2022OCT,Liu2024TACO,He2025GazeGuided,Cho2014GRU}.
These limitations motivate a hierarchical approach to explicitly address semantic and geometric ambiguity in coordinated hand-object interaction forecasting.

\begin{sidewaystable*}[p]
    \centering
    \caption{Comparison of dataset scale, annotations, and supported prediction tasks.}
    \label{tab1}
    \begingroup
    \ifdefined\hypersetup\hypersetup{hidelinks}\fi
    \small
    \newcommand{\gapDatasetSamples}{6pt}
    \newcommand{\gapSamplesObjects}{5.5pt}
    \newcommand{\gapObjectsIntents}{5.5pt}
    \newcommand{\gapIntentsGaze}{6pt}
    \newcommand{\gapGazeHand}{6pt}
    \newcommand{\gapHandObject}{6pt}
    \newcommand{\gapObjectSemantics}{0pt}
    \newcommand{\gapSemanticsScene}{3pt}
    \newcommand{\gapSceneIntention}{7pt}
    \newcommand{\gapIntentionMotion}{4pt}
    \newcommand{\gapMotionJoint}{9pt}
    \newcommand{\groupRuleInset}{1pt}
    \setlength{\tabcolsep}{0pt}
    \renewcommand{\arraystretch}{1.08}
    \setlength{\aboverulesep}{0.3ex}
    \setlength{\belowrulesep}{0.3ex}
    \newcommand{\tblHandTitle}{Hand repr.}
    \newcommand{\tblObjectTitle}{Object pose}
    \newcommand{\tblTaskTitle}[1]{\tblHead{#1}}
    \newcommand{\tblDatasetRows}{%
        \toprule
        \multirow{2}{*}{\tblHead{\textbf{Dataset}}} &
        \multicolumn{3}{c}{\hspace*{\gapDatasetSamples}\tblHead{\textbf{Dataset scale}}} &
        \multicolumn{5}{c}{\hspace*{\gapIntentsGaze}\tblHead{\textbf{Annotations}}} &
        \multicolumn{3}{c}{\hspace*{\gapSceneIntention}\tblHead{\textbf{Prediction support}}} \\
        \cmidrule(l{\dimexpr\gapDatasetSamples+\groupRuleInset\relax}r{\groupRuleInset}){2-4}
        \cmidrule(l{\dimexpr\gapIntentsGaze+\groupRuleInset\relax}r{\groupRuleInset}){5-9}
        \cmidrule(l{\dimexpr\gapSceneIntention+\groupRuleInset\relax}r{\groupRuleInset}){10-12}
        & \tblHead{\#Samples} & \tblHead{\#Objects} & \tblHead{\#Intents} &
        \tblHead{Gaze} & \tblHead{\tblHandTitle} & \tblHead{\tblObjectTitle} & \tblHead{Semantics} & \tblHead{Scene geometry} &
        \tblTaskTitle{Intention} & \tblTaskTitle{Motion} & \tblTaskTitle{Joint} \\
        \midrule
        Ego4D~\citep{Grauman2022Ego4D} & 9,645$^{r}$ & --- & --- & \tblYes\tblMark{*} & 2D boxes & --- & Action & Meshes & \tblNo & \tblYes & \tblNo \\
        EgoThink~\citep{Cheng2024EgoThink} & 700$^{i}$ & --- & --- & \tblNo & --- & --- & Visual QA & --- & \tblYes & \tblNo & \tblNo \\
        EgoPAT3D~\citep{Li2022EgoPAT3D} & 150$^{r}$ & --- & --- & \tblNo & 3D point & --- & --- & Point clouds & \tblYes\tblMark{\dagger} & \tblYes\tblMark{\ddagger} & \tblNo \\
        EgoGazeVQA~\citep{Peng2025EgoGazeVQA} & 913$^{c}$ & --- & --- & \tblYes & --- & --- & Intent QA & --- & \tblYes & \tblNo & \tblNo \\
        GazeHOI~\citep{Tian2026GazeHOI} & 1,378$^{c}$ & 33 & --- & \tblYes & MANO & 6-DoF & Task text & --- & \tblNo & \tblYes & \tblNo \\
        HOT3D~\citep{Banerjee2025HOT3D} & 424$^{r}$ & 33 & --- & \tblYes\tblMark{*} & MANO & 6-DoF & --- & Point clouds & \tblNo & \tblYes & \tblNo \\
        H2O~\citep{Kwon2021H2O} & 933$^{c}$ & 8 & --- & \tblNo & MANO & 6-DoF & Action & Depth maps & \tblNo & \tblYes\tblMark{\ddagger} & \tblNo \\
        TACO~\citep{Liu2024TACO} & 2.5K$^{s}$ & 196 & --- & \tblNo & MANO & 6-DoF & Action & Depth maps & \tblNo & \tblYes & \tblNo \\
        \midrule
        \rowcolor{EgoAIMRow}
        \textbf{EgoAIM (ours)} & \textbf{7,429}$^{w}$ & \textbf{33} & \textbf{9,038} & \tblYes & MANO & 6-DoF & \textbf{Per-hand intent} & \textbf{Support planes} & \tblYes & \tblYes & \tblYes \\
        \bottomrule
    }
    \newcommand{\tblDatasetTabular}{%
    \begin{tabular}{>{\footnotesize}l
        >{\footnotesize\hspace*{\gapDatasetSamples}}c
        >{\footnotesize\hspace*{\gapSamplesObjects}}c
        >{\footnotesize\hspace*{\gapObjectsIntents}}c
        >{\footnotesize\hspace*{\gapIntentsGaze}}c
        >{\footnotesize\hspace*{\gapGazeHand}}c
        >{\footnotesize\hspace*{\gapHandObject}}c
        >{\footnotesize\hspace*{\gapObjectSemantics}}c
        >{\footnotesize\hspace*{\gapSemanticsScene}}c
        >{\footnotesize\hspace*{\gapSceneIntention}}c
        >{\footnotesize\hspace*{\gapIntentionMotion}}c
        >{\footnotesize\hspace*{\gapMotionJoint}}c}
        \tblDatasetRows
    \end{tabular}%
    }
    \sbox{\tblMeasureBox}{\tblDatasetTabular}
    \typeout{TABLE WIDTH: \the\wd\tblMeasureBox}
    \typeout{AVAILABLE WIDTH: \the\linewidth}
    \ifdim\wd\tblMeasureBox>\textwidth
        \PackageWarning{EgoAIM}{Table exceeds text width; reduce the manual gap settings}
    \fi
    \usebox{\tblMeasureBox}
    \par\vspace{6pt}
    \begin{minipage}{\textwidth}
        \linespread{1}\fontsize{9pt}{12pt}\selectfont
        Units: $r$ recordings/videos, $s$ sequences, $c$ clips, $i$ images, $w$ windows (counts not directly comparable).
        \#Objects: instances; \#Intents: non-null per-hand annotations. ---: unreported/inapplicable; \tblNo: unavailable/not directly supported.
        $^{*}$Gaze subset; $^{\dagger}$target localization; $^{\ddagger}$derived trajectory protocols~\citep{Bao2023USST}.
        Intention includes QA; Joint requires established paired intention and articulated 3D HOI motion evaluation on shared future windows.
        Scene geometry excludes object meshes.
    \end{minipage}
    \endgroup
\end{sidewaystable*}

\subsection{Conditional Hand-Object Motion Generation}
Extensive research has leveraged diverse conditioning modalities to drive hand-object motion generation~\citep{Sui2026InteractionMotionSurvey,Tevet2023MDM,Hu2026A3ffordHOI}.
Language-conditioned methods provide semantic control over hand articulation and object motion, with DiffH2O separating grasping from manipulation while maintaining continuity between the stages~\citep{Christen2024DiffH2O}.
Since semantic descriptions typically leave precise surface contacts underdetermined, Text2HOI predicts contact maps from text and object geometry to guide joint hand-object motion diffusion~\citep{Cha2024Text2HOI}.
Contact representations also support generation conditioned on prescribed object trajectories: BimArt generates temporally varying contact maps to guide bimanual motion for articulated objects~\citep{Zhang2025BimArt}.
For interactions involving multiple entities, SyncDiff jointly models absolute and relative motions, explicitly synchronizing their generation under action and geometry conditions~\citep{He2025SyncDiff}.
GHO-Diffusion further explores gaze-conditioned synthesis, generating object motion followed by corresponding hand motion using gaze sequences spanning the synthesis interval~\citep{Tian2026GazeHOI}.
However, its gaze-motion consistency formulation does not explicitly model the temporal lead of gaze, with fixations often preceding task-relevant hand actions~\citep{Johansson2001EyeHandCoordination}.
Furthermore, its reliance on future gaze observations within the generated interval precludes its direct application to forecasting based solely on historical observations.
Even when egocentric visual context is available, LatentAct requires an action description and a specified 3D contact point~\citep{Prakash2025LatentAct}, while MEgoHand relies on action instructions and an initial hand pose to guide motion generation~\citep{Zhou2025MEgoHand}.
These methods primarily address how an interaction should be realized under supplied guidance.
To support seamless interaction in practice, systems must autonomously infer the user's intention from ongoing observations and translate it into conditions for motion generation.

\subsection{Gaze-Aware Intention Understanding}
Gaze-aware intention understanding draws on behavioral evidence that gaze can precede task-relevant hand-object actions~\citep{Johansson2001EyeHandCoordination}.
GIMO integrates gaze through bidirectional attention~\citep{Zheng2022GIMO}, while GazeMotion forecasts gaze from its history to guide motion prediction~\citep{Hu2024GazeMotion}.
These methods use gaze to improve motion prediction, without producing explicit object-action intention descriptions.
For semantic interpretation, GLAM aligns gaze scanpaths with language, encoding fixation location, duration, and order to identify visual-search targets~\citep{Mondal2025GLAM}.
However, its predictions only identify targets, without specifying how the user intends to interact with the object.
Advances in VLMs have enabled richer scene understanding~\citep{Qiu2026EgoPlanBench2}, yet effective gaze integration remains challenging.
Peng et al. explore input-level prompting through textual gaze coordinates, visual markers, and saliency maps~\citep{Peng2025EgoGazeVQA}.
The visual-marker variant, however, alters scene appearance and may bias semantic interpretation by mixing gaze annotations with scene content.
Beyond input-level prompting, Voila-A injects gaze heatmaps into a Perceiver resampler's attention keys, but its heatmap encoding does not explicitly separate current fixation from gaze history~\citep{Yan2024Voila-A}.
Gaze-VLM regularizes attention using temporally aggregated gaze during training~\citep{Pani2025Gaze-VLM}, but its video-only inference cannot directly exploit observed gaze to distinguish plausible intentions.
Gaze2HOI instead represents current fixation and gaze history as separate, scene-aligned signals that drive gated scale-and-shift modulation of VLM visual features at multiple levels.
This design conditions structured future intention inference on evolving attention, providing semantic guidance for geometric grounding and coordinated hand-object motion forecasting.

\section{EgoAIM Benchmark}
This section introduces EgoAIM, a phase-aware benchmark for Egocentric Anticipation of Intention and Motion.
As summarized in Table~\ref{tab1}, existing benchmarks provide semantic and motion annotations, yet offer limited support for joint evaluation of fine-grained future intentions and coordinated 3D hand-object motion.
EgoAIM bridges this gap by linking per-hand future intentions to continuous 3D kinematics across four interaction phases.
Support-plane and placement-target annotations further characterize the environmental geometry relevant to object support and placement.

\begin{figure*}[!t]
    \centering
    \includegraphics[width=0.65\textwidth]{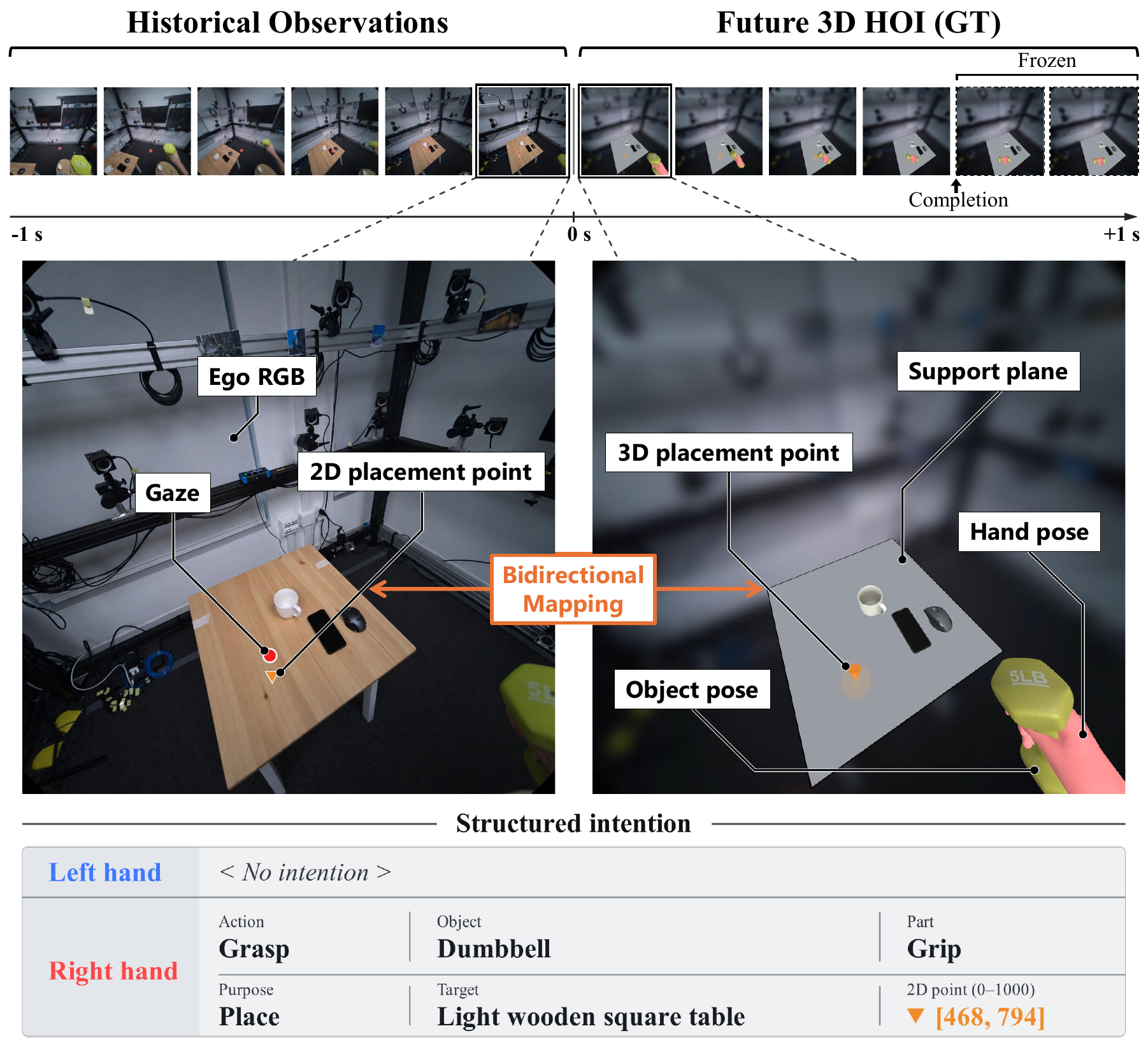}
    \caption{\textbf{\emph{Illustration of EgoAIM annotations.}} (Top) A 2-second window pairing egocentric observations with future hand-object interaction. (Middle) Alignment between 2D egocentric observations and 3D geometric annotations. (Bottom) Structured future intentions annotated separately for each hand.}
    \label{fig3}
    \vspace{2 mm}
\end{figure*}

\subsection{Benchmark Construction}
EgoAIM integrates egocentric RGB videos, eye gaze, articulated hand poses, and 6-DoF object trajectories sourced from HOT3D-Aria~\citep{Banerjee2025HOT3D}, all temporally aligned with the RGB timeline.
To capture complete intended interactions, we extract 60-frame candidate windows with a 5-frame stride, providing alternative temporal boundaries around interaction events and reducing the risk of intention truncation.
At 30 fps, each window comprises one second of historical observations and one second of future motion, as illustrated in Fig.~\ref{fig3}.
All candidate windows undergo manual review to exclude those with incomplete interactions, missing hand tracking, and interrupted tracking of interacting objects.
For the \emph{Establish Contact} and \emph{Put Down} phases, we additionally identify the completion of the annotated interaction.
When completion precedes the end of the prediction horizon, the final valid hand-object state is frozen for the remaining frames, preventing subsequent unannotated actions from contaminating the motion target.
This process yields 7,429 forecasting windows covering 33 object instances, with 9,038 non-null per-hand intention annotations.

We distinguish \textbf{four phases} according to evolving hand-object relations, reflecting distinct intention inference and motion coordination demands throughout the interaction process.
\textbf{\emph{Establish Contact}} involves reaching toward an object and establishing a grasp or touch, requiring anticipation of the intended target and contact configuration.
\textbf{\emph{Move}} concerns transporting an object while maintaining contact, emphasizing coordinated hand-object motion.
\textbf{\emph{Manipulate}} covers functional operations such as stirring or pressing, where task-specific articulation and inter-object relationships become important.
\textbf{\emph{Put Down}} involves placing an object at an intended destination, introducing constraints from the target location and supporting surface.
This organization enables phase-specific evaluation of intention inference and motion forecasting, revealing performance differences that aggregate metrics may obscure.
We manually assign a phase label to each window and annotate intentions separately for each hand to capture the potentially different roles of the two hands within the interaction.

\begin{figure*}[!t]
    \centering
    \includegraphics[width=\textwidth]{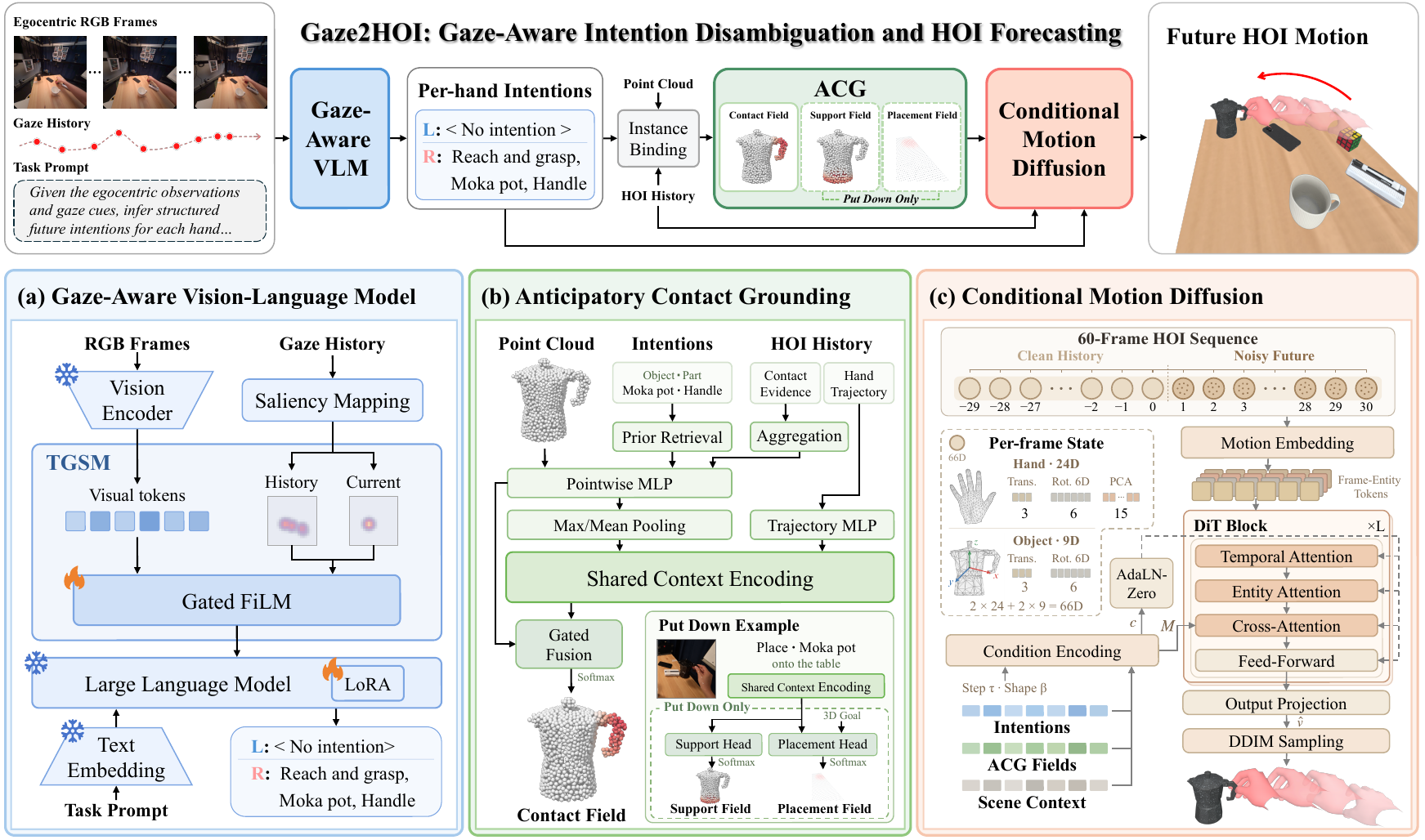}
    \caption{\emph{\textbf{Overview of Gaze2HOI.}}
    Hierarchical inference from egocentric observations to future hand-object interactions:
    (a) TGSM modulates VLM visual features with gaze cues to infer structured per-hand intentions.
    (b) ACG grounds inferred intentions into explicit 3D contact, support, and placement fields.
    (c) Conditional motion diffusion leverages semantic and geometric conditions to jointly forecast articulated hand motion and 6-DoF object trajectories.}
    \label{fig4}
    \vspace{-2 mm}
\end{figure*}

\subsection{Intention and Scene Annotations}
To capture the asymmetric nature of bimanual manipulation, we annotate fine-grained future intentions separately for each hand using a \textbf{structured intention representation}.
This representation explicitly couples physical interactions with functional goals: the \emph{Action}, \emph{Object}, and \emph{Part} fields describe how the hand engages with an object, while \emph{Purpose} and \emph{Target} identify the intended operation and its target object or surface.
For placement actions, an additional \emph{2D Point} field localizes the intended destination in the final observed RGB frame.
Calibrated camera projection and ray-plane intersection with the annotated support planes establish a bidirectional mapping between this image point and its 3D placement target.
Within this representation, hands without task-directed involvement receive a \emph{No intention} label, whereas active support and stabilization are annotated explicitly.
A controlled vocabulary and object-specific part definitions standardize annotation terminology across samples.
To ensure annotation fidelity, intention labels are cross-checked against observed and future video frames, rendered 3D hand-object motion, and per-hand action timelines.

Beyond structured intention labels, EgoAIM provides \textbf{scene annotations} specifying the shapes, spatial extents, and poses of furniture support surfaces.
We first derive surface heights from the mesh geometry and poses of tracked objects resting on the furniture.
Surface boundaries are then manually delineated in multiple RGB frames and back-projected onto the corresponding horizontal planes using calibrated camera parameters to recover surface shapes and dimensions.
Subsequently, we position and orient the reconstructed surfaces in each sequence's gravity-aligned coordinate system and manually refine them through reprojection across frames.
Each surface is associated with a furniture label and registered alongside the hand and object trajectories, providing a common spatial reference for relating hand-object motion to the environment.

\section{Gaze2HOI Framework}
This section presents the Gaze2HOI framework for forecasting coordinated hand-object interactions from egocentric observations.
We first formulate the task and then describe how the gaze-aware vision-language model and Anticipatory Contact Grounding provide semantic and geometric conditions for motion diffusion.

\subsection{Problem Formulation}
Given a history of egocentric observations together with scene information, Gaze2HOI infers structured future intentions for each hand and jointly forecasts articulated hand motion and the 6-DoF trajectories of interacting objects. 
The overall forecasting task is formulated as
\begin{equation}
(\hat{\mathcal I},\hat{\mathcal X}_{\mathrm{fut}})
=
f\!\left(
\mathcal V_{\mathrm{obs}},
\mathcal G_{\mathrm{obs}},
\mathcal X_{\mathrm{obs}},
\mathcal P,
\mathcal S
\right),
\label{eq1}
\end{equation}
where $f$ denotes the overall prediction pipeline.
$\mathcal V_{\mathrm{obs}}$, $\mathcal G_{\mathrm{obs}}$, and $\mathcal X_{\mathrm{obs}}$ represent observed RGB frames, gaze cues, and 3D hand-object states, respectively.
The task prompt $\mathcal P$ contains forecasting instructions, scene-specific descriptions, and output requirements.
Scene information $\mathcal S$ includes object point clouds and support surfaces.
The outputs are structured per-hand intentions $\hat{\mathcal I}$ and a future state sequence $\hat{\mathcal X}_{\mathrm{fut}}\in\mathbb R^{T_{\mathrm{fut}}\times66}$.
Each state is formed by concatenating two 24D hand states (comprising wrist translation, 6D rotation, and 15 MANO~\citep{Romero2017MANO} PCA pose coefficients) with two 9D object states (comprising translation and 6D rotation).
Following EgoAIM, we observe $T_{\mathrm{obs}}=30$ frames and predict $T_{\mathrm{fut}}=30$ frames at 30 fps.

As illustrated in Fig.~\ref{fig4}, Gaze2HOI first performs \textbf{semantic disambiguation} by inferring structured per-hand intentions $\hat{\mathcal I}=f_{\phi}(\mathcal V_{\mathrm{obs}},\mathcal G_{\mathrm{obs}},\mathcal P)$.
Here, $\hat{\mathcal I}=(\hat{\mathcal I}^{L},\hat{\mathcal I}^{R})$ specifies each hand's intended action, object, part, purpose, and target, with an additional 2D destination for \emph{Put Down}.
The inferred object identities establish the hand-object bindings, yielding the updated observation sequence $\widetilde{\mathcal X}_{\mathrm{obs}}$.
Next, the ACG module performs \textbf{geometric disambiguation} by combining the inferred intentions, object geometry, and observed interactions to predict contact, support, and placement fields, collectively denoted by $\hat{\mathcal F}_{\mathrm{ACG}}$.
Finally, leveraging these semantic and geometric predictions, the conditional diffusion generator $f_{\omega}$ jointly forecasts future hand-object motion as $\hat{\mathcal X}_{\mathrm{fut}}=f_{\omega}(\widetilde{\mathcal X}_{\mathrm{obs}},\hat{\mathcal I},\hat{\mathcal F}_{\mathrm{ACG}},\mathcal S)$.

\par
\vspace{6 mm}

\subsection{Gaze-Aware Vision-Language Model}
Egocentric observations often support multiple plausible interactions before the user's intention becomes evident from hand kinematics, reflecting semantic ambiguity in intention inference.
While gaze provides anticipatory cues for object manipulation~\citep{Johansson2001EyeHandCoordination}, individual gaze samples vary in stability and temporal relevance.
Uniform aggregation fails to explicitly distinguish stable fixations from rapid gaze shifts or capture their evolving relevance as interactions progress.
To support semantic disambiguation, we augment the Qwen3-VL~\citep{Bai2025Qwen3VL} backbone with Temporal Gaze Saliency Modulation (TGSM), which encodes current and historical gaze into distinct spatial channels and applies gated feature-wise modulation to the corresponding visual tokens.

For each input RGB frame at time $t$, we reproject the available 3D gaze points from the observation window into its image coordinates using the corresponding camera pose.
Let $\overline{\mathbf g}_{t'\rightarrow t}$ denote the projected gaze location from time $t'$, and $\mathcal T_t^{\mathrm{valid}}$ represent the set of valid gaze sample indices up to and including $t$.
Using a Gaussian kernel $K_{\sigma_{\mathrm g}}(\mathbf d)=\exp(-\|\mathbf d\|_2^2/(2\sigma_{\mathrm g}^2))$, we construct historical and current saliency maps:
\begin{gather}
A_t^{\mathrm{hist}}(\mathbf u)
=
\max_{t'\in\mathcal T_t^{\mathrm{valid}}}
\left[
w_{t',t}
K_{\sigma_{\mathrm g}}
\!\left(\mathbf u-\overline{\mathbf g}_{t'\rightarrow t}\right)
\right],
\label{eq2}\\
A_t^{\mathrm{cur}}(\mathbf u)
=
K_{\sigma_{\mathrm g}}
\!\left(\mathbf u-\overline{\mathbf g}_{t\rightarrow t}\right),
\label{eq3}
\end{gather}
where $\mathbf u$ denotes a grid coordinate and $\sigma_{\mathrm g}$ controls the kernel's spatial spread.
The weight $w_{t',t}=q_{t',t}\exp(-\lambda_{\mathrm{time}}\Delta_{t',t})$ combines fixation strength and recency to emphasize stable, recent gaze cues.
We derive the fixation weight $q_{t',t}$ from the complementary empirical cumulative distribution function (CDF) of projected gaze speed with a positive floor, assigning greater weight to stable fixations.
The temporal decay reduces the influence of older gaze samples, with $\Delta_{t',t}$ denoting elapsed time in seconds and $\lambda_{\mathrm{time}}>0$ controlling the decay rate.
To handle missing data, we set the current map to zero when the current gaze point is invalid, and the historical map to zero when no valid samples are available.
Together, these two channels encode both the instantaneous gaze focus and its spatially aligned, temporally weighted historical context.

To condition the visual features, we spatially align the two saliency maps with the visual token grid via adaptive average pooling, yielding a two-channel gaze descriptor $\mathbf a\in\mathbb R^2$ for each visual token $\mathbf e\in\mathbb R^{d_{\mathrm v}}$.
Two bias-free MLPs, $\boldsymbol\gamma_{\phi}$ and $\mathbf b_{\phi}$, transform this descriptor into channel-wise scale and shift vectors, allowing spatially corresponding gaze cues to adjust each token's visual representation.
The resulting gated feature-wise modulation is defined as follows:
\begin{equation}
\begin{gathered}
\widetilde{\mathbf e}
=
\mathbf e\odot
\bigl[
\mathbf 1+
\lambda_{\mathrm{mod}}
\tanh(\rho_{\mathrm{scale}})
\tanh\!\bigl(\boldsymbol\gamma_{\phi}(\mathbf a)\bigr)
\bigr]
\\[2pt]
{}+
\tanh(\rho_{\mathrm{shift}})
\,\mathbf b_{\phi}(\mathbf a),
\end{gathered}
\label{eq4}
\end{equation}
where $\odot$ denotes element-wise multiplication, and $\rho_{\mathrm{scale}}$ and $\rho_{\mathrm{shift}}$ are learnable scalar gates controlling the strength of scale and shift modulation, respectively.
The coefficient $\lambda_{\mathrm{mod}}=0.1$ bounds the multiplicative deviation from the original features.
The projection networks are shared between the final visual embeddings and the intermediate visual features supplied to the language model.
With bias-free projections, a zero gaze descriptor leaves the original visual token unchanged.

Conditioned on the gaze-modulated visual representations $\widetilde{\mathbf E}$ and task prompt $\mathcal P$, the language model autoregressively predicts structured per-hand intentions.
For a serialized ground-truth response $\mathbf y=(y_1,\ldots,y_{N_{\mathrm{ans}}})$, the corresponding training objective is defined as
\begin{equation}
\mathcal L_{\mathrm{intent}}
=
-\frac{1}{N_{\mathrm{ans}}}
\sum_{m=1}^{N_{\mathrm{ans}}}
\log p_{\phi}\!\left(
y_m
\mid
\mathbf y_{<m},
\widetilde{\mathbf E},
\mathcal P
\right),
\label{eq5}
\end{equation}
where $N_{\mathrm{ans}}$ is the number of supervised response tokens and $\mathbf y_{<m}$ denotes the ground-truth prefix used for teacher forcing.
To preserve pretrained knowledge, we freeze the backbone weights and optimize only the TGSM modules alongside low-rank adaptation (LoRA)~\citep{Hu2022LoRA} parameters in the language model.
At inference, the decoded response is parsed into structured per-hand intentions $\hat{\mathcal I}$, facilitating \textbf{semantic disambiguation} and providing explicit guidance for downstream object instance binding and anticipatory contact grounding.

\subsection{Anticipatory Contact Grounding}
While structured intentions identify the target object and active part, they leave the precise spatial contact configuration underdetermined.
To support \textbf{geometric disambiguation}, Anticipatory Contact Grounding (ACG) integrates semantic priors, observed contact, and hand trajectories to predict future contact distributions.
For each valid hand-object binding, a contact prior $\boldsymbol\pi_{\mathrm{prior}}$ is retrieved from a precomputed contact dictionary using the inferred object and part identities and the hand side.
We represent all contact distributions on a fixed set of surface points in the bound object's local coordinate system (omitting the hand index below for clarity).
To construct the complementary observed contact evidence $\boldsymbol\pi_{\mathrm{obs}}$, we convert distances from object surface points to the hand mesh into normalized contact maps using a truncated Gaussian kernel at each valid contact frame.
These maps are averaged over observed contact frames, with zero evidence assigned when no valid contact is available.
Functionally, the semantic prior provides anticipatory spatial guidance even before physical contact occurs, while the observed evidence grounds the prediction in the ongoing interaction.

We first encode each surface point via a pointwise MLP using its coordinates, density descriptors of the prior and observed contact, along with its distance to the last valid wrist position.
After global maximum and average pooling, these summarized features are combined with an MLP-encoded wrist trajectory and source reliability and validity cues to form a shared context representation.
This shared context determines the fusion weights, while its combination with local point features produces a pointwise correction $\delta_i$.
The future contact field is obtained through gated log-linear fusion:
\begin{gather}
\ell_i
=
w_{\mathrm{prior}}\log\pi_{\mathrm{prior},i}
+
w_{\mathrm{obs}}\log\pi_{\mathrm{obs},i}
+
\delta_i,
\label{eq6}\\[3pt]
\hat{\boldsymbol\pi}_{\mathrm{contact}}
=
\operatorname{softmax}(\boldsymbol\ell),
\label{eq7}
\end{gather}
where $i$ indexes object surface points.
The independent gates $w_{\mathrm{prior}},w_{\mathrm{obs}}\geq0$ adjust each source's contribution for the current interaction, while $\delta_i$ provides a local correction conditioned on geometry and the observed trajectory.
Following log-linear fusion, softmax converts the resulting logits into a contact probability distribution.

For \emph{Put Down}, ACG further predicts the object regions expected to contact the supporting surface and their spatial footprint around the placement goal.
The support head first refines an object-specific prior $\boldsymbol\pi_{\mathrm{sup}}^{\mathrm{prior}}$ (estimated from training placements) using the shared context alongside local geometric and contact cues.
With a context-dependent gate $w_{\mathrm{sup}}\geq0$ and pointwise corrections $\boldsymbol\delta_{\mathrm{sup}}$, the support field is computed as $\hat{\boldsymbol\pi}_{\mathrm{sup}}=\operatorname{softmax}(w_{\mathrm{sup}}\log\boldsymbol\pi_{\mathrm{sup}}^{\mathrm{prior}}+\boldsymbol\delta_{\mathrm{sup}})$, where softmax normalizes over object surface points.
To ground the spatial footprint, we lift the predicted 2D destination to a 3D goal by intersecting its camera ray with a scene support plane, establishing the center of a surface-aligned local grid.
The placement head forms $\mathbf z_{\mathrm{plc}}$ from shared context, pooled support features, and surface-relative object heading.
Finally, a convolutional decoder $D_{\psi}$ processes this representation to yield the placement field $\hat{\boldsymbol\pi}_{\mathrm{plc}}=\operatorname{softmax}(D_{\psi}(\mathbf z_{\mathrm{plc}}))$, which is normalized over the local grid cells.

ACG is trained with annotated intentions and geometric targets derived from future interactions.
We construct the contact target by averaging normalized contact maps over valid future contact frames selected according to the interaction phase.
For \emph{Put Down}, the support target normalizes truncated Gaussian weights computed from surface-point distances to the support plane at the terminal object pose.
The placement target is obtained by projecting these points onto the local grid, accumulating their support weights with Gaussian kernels, and normalizing the result.
For the field types $\mathcal J=\{\mathrm{contact},\mathrm{sup},\mathrm{plc}\}$, the training objective is
\begin{equation}
\mathcal L_{\mathrm{ACG}}
=
\sum_{j\in\mathcal J}
\lambda_j
\bigl\langle
D_{\mathrm{KL}}(
\boldsymbol\pi_j^{*}
\Vert
\hat{\boldsymbol\pi}_j
)
\bigr\rangle_j,
\label{eq8}
\end{equation}
where $\boldsymbol\pi_j^{*}$ denotes the target distribution and $\lambda_j$ weights the corresponding loss.
The operator $\langle\cdot\rangle_j$ averages over samples with valid supervision for field $j$ in the minibatch, with zero contribution when none are available.
Support and placement losses apply only to valid \emph{Put Down} examples.
After training, the frozen ACG predicts geometric fields $\hat{\mathcal F}_{\mathrm{ACG}}$, which mitigate geometric ambiguity and provide explicit spatial conditioning for the subsequent motion diffusion model.

\subsection{Conditional Motion Diffusion}
Structured intentions and anticipatory contact fields constrain the semantics and geometry of an interaction, but do not uniquely determine its temporal evolution.
To model this remaining uncertainty, we employ a conditional motion diffusion model to jointly generate articulated hand motion and object trajectories.
The forecasting task is formulated as temporal inpainting: the observed hand-object states remain clean, while only future states are perturbed.
Let $\mathcal X_{\mathrm{fut}}^{(0)}$ denote the normalized clean future sequence.
At diffusion step $\tau$, the noisy future states and the corresponding $v$-prediction target~\citep{Salimans2022ProgressiveDistillation} are defined as follows:
\begin{gather}
\mathcal X_{\mathrm{fut}}^{(\tau)}
=
\alpha_\tau\mathcal X_{\mathrm{fut}}^{(0)}
+
\sigma_\tau\boldsymbol\epsilon,
\label{eq9}\\[3pt]
\mathbf v_{\mathrm{fut}}^{(\tau)}
=
\alpha_\tau\boldsymbol\epsilon
-
\sigma_\tau\mathcal X_{\mathrm{fut}}^{(0)},
\label{eq10}
\end{gather}
where $\boldsymbol\epsilon$ is standard Gaussian noise, and the noise-schedule coefficients satisfy \mbox{$\alpha_\tau^2+\sigma_\tau^2=1$}.
The denoiser receives the clean history concatenated with the noisy future, allowing observed interactions to constrain the entire predicted sequence.

Semantic and geometric guidance is incorporated through two complementary mechanisms: a token memory $\mathbf M$ provides structured cues through cross-attention, while a global conditioning vector $\mathbf c$ modulates the denoising sublayers.
The memory integrates per-hand intention attributes with scene and goal information.
To preserve the spatial structure of ACG predictions, we partition each bound object's surface into local regions and construct region tokens from local geometry and aggregated contact and support probability mass.
These region tokens supplement the semantic and scene representations in $\mathbf M$.
In parallel, an MLP combines embeddings of the diffusion step $\tau$ and MANO shape parameters $\boldsymbol\beta$ with pooled intention and goal representations to produce the global conditioning vector $\mathbf{c}$.

To jointly model temporal evolution and hand-object coordination, the denoiser represents each frame using four entity tokens corresponding to the two hands and their associated objects.
Raw states are mapped into a shared latent space and augmented with embeddings indicating frame position, entity identity, and observation validity.
Each DiT block processes these tokens through temporal attention (across frames per entity) and intra-frame entity attention (among the four tokens), followed by cross-attention~\citep{Vaswani2017Transformer} to $\mathbf M$ and a feed-forward network.
For input tokens $\mathbf Y$, each residual sublayer performs the AdaLN-Zero~\citep{Peebles2023DiT} modulated update $\mathbf Y^{+} = \mathbf Y + \boldsymbol\rho_{\mathbf c} \odot \mathcal A\bigl( (\mathbf 1+\boldsymbol\gamma_{\mathbf c}) \odot \operatorname{LN}(\mathbf Y) + \mathbf b_{\mathbf c} \bigr)$, where $\operatorname{LN}$ is layer normalization and $\mathcal A$ denotes the respective attention or feed-forward operation.
The scale, shift, and residual-gate vectors $\{\boldsymbol\gamma_{\mathbf c}, \mathbf b_{\mathbf c}, \boldsymbol\rho_{\mathbf c}\}$ are generated from $\mathbf c$ separately for each sublayer and shared across tokens.
The modulation projection is initialized to zero, making each residual sublayer an identity mapping at initialization.
After $L$ blocks and a final adaptive normalization, output projections yield the joint $v$-prediction, with future component $\hat{\mathbf v}_{\mathrm{fut}}^{(\tau)}$.

During motion training, annotated intentions and fields predicted by the frozen ACG model provide semantic and geometric conditions.
The denoising objective penalizes discrepancies between predicted and target $v$ values, with supervision restricted to valid future state components:
\begin{equation}
\mathcal L_v
=
\frac{
\left\|
\mathbf W_{\mathrm{fut}}\odot
\left(
\hat{\mathbf v}_{\mathrm{fut}}^{(\tau)}
-
\mathbf v_{\mathrm{fut}}^{(\tau)}
\right)
\right\|_F^2
}{
\max\!\left(1,\|\mathbf W_{\mathrm{fut}}\|_1\right)
},
\label{eq11}
\end{equation}
where $\mathbf W_{\mathrm{fut}}$ is a binary mask identifying valid future state components across the minibatch, and $\|\cdot\|_F$ and $\|\cdot\|_1$ denote the Frobenius and entrywise $L_1$ norms, respectively.
To encourage coordinated and physically plausible interactions, we apply auxiliary losses to denormalized clean-state estimates recovered from the predicted $v$.
The overall training objective is defined as follows:
\begin{equation}
\begin{aligned}
\mathcal L_{\mathrm{motion}}
={}&
\mathcal L_v
+\mathcal L_{\mathrm{geo}}
+\mathcal L_{\mathrm{rel}}
+\mathcal L_{\mathrm{contact}}
\\[3pt]
&+
\mathcal L_{\mathrm{obj}}
+\mathcal L_{\mathrm{sup}}
+\mathcal L_{\mathrm{move}},
\end{aligned}
\label{eq12}
\end{equation}
where $\mathcal L_{\mathrm{geo}}$ supervises translations, velocities, and rotations, while $\mathcal L_{\mathrm{rel}}$ constrains relative hand-object motion by matching wrist-relative poses and their temporal changes.
Contact consistency and collision avoidance are encouraged through $\mathcal L_{\mathrm{contact}}$, which combines regional contact-distance matching with hand-object penetration penalties, and $\mathcal L_{\mathrm{obj}}$, which penalizes object-object penetration.
The support term $\mathcal L_{\mathrm{sup}}$ addresses furniture penetration, terminal placement, and placement-goal errors, while $\mathcal L_{\mathrm{move}}$ supervises object transport relative to the intended target.
Each auxiliary term uses its corresponding validity mask.
To avoid imposing interaction constraints on unreliable clean-state estimates, the interaction losses are activated only when $\alpha_\tau^2\geq0.25$, whereas $\mathcal L_{\mathrm{geo}}$ is evaluated at all noise levels.

Intention and goal conditions are randomly dropped during training to enable classifier-free guidance (CFG)~\citep{Ho2022CFG}.
At inference, DDIM sampling~\citep{Song2021DDIM} with CFG generates future motion conditioned on VLM-predicted intentions and the corresponding ACG fields.
Future states are initialized with Gaussian noise, while observed states are restored after each sampling step.
The resulting future sequence is denormalized to recover articulated hand motion and 6-DoF object trajectories.
Ultimately, by coupling structured semantic intentions and explicit geometric grounding with stochastic motion generation, our unified framework substantially mitigates semantic and geometric ambiguity, yielding physically plausible and temporally coherent interaction forecasts.

\begin{table*}[!t]
    \centering
    \caption{Quantitative comparison of hand-object motion forecasting on EgoAIM.}
    \label{tab2}
    \begingroup
    \ifdefined\hypersetup\hypersetup{hidelinks}\fi
    \linespread{1}\fontsize{8.2pt}{11pt}\selectfont
    \newcommand{\gapMethodHand}{2pt}
    \newcommand{\gapHandObject}{4pt}
    \newcommand{\gapObjectPen}{1.5pt}
    \newcommand{\gapPenDivider}{1.5pt}
    \newcommand{\gapDividerReach}{1.5pt}
    \newcommand{\gapReachGoal}{1.5pt}
    \newcommand{\gapGoalManipulate}{1.5pt}
    \newcommand{\gapManipulatePlace}{1.5pt}
    \newcommand{\forecastRuleInset}{1pt}
    \setlength{\tabcolsep}{0pt}
    \renewcommand{\arraystretch}{1.08}
    \setlength{\aboverulesep}{0.2ex}
    \setlength{\belowrulesep}{0.2ex}
    \setlength{\arrayrulewidth}{0.35pt}
    \definecolor{ForecastOursRow}{RGB}{242,242,242}
    \newcommand{\forecastHead}[1]{\mbox{\scalebox{0.85}{\fontsize{8.5}{10}\selectfont #1}}}
    \newcommand{\forecastMetricHead}[2]{%
        \forecastHead{#1\,$#2$}%
    }
    \newcommand{\forecastPairSep}{\hspace{0.6pt}/\hspace{0.6pt}}
    \newcommand{\forecastOursCell}[1]{%
        \begingroup
        \setbox0=\hbox{\textbf{92.43}}%
        \raisebox{\dimexpr(\ht\csname @arstrutbox\endcsname-\dp\csname @arstrutbox\endcsname-\ht0+\dp0)/2\relax}[0pt][0pt]{#1}%
        \endgroup
    }
    \newcommand{\forecastMethod}[2]{#1\,{\fontsize{7pt}{8pt}\selectfont\scalebox{0.85}{\citep{#2}}}}
    \newcommand{\forecastNA}{---}
    \newcommand{\forecastDivider}{{\color{black!45}\vrule width\arrayrulewidth}}
    \newcommand{\forecastHeaderDivider}{{%
        \color{black!45}%
        \rlap{\raisebox{-\dimexpr\dp\csname @arstrutbox\endcsname+\aboverulesep+\cmidrulewidth+\belowrulesep+0.1pt\relax}[0pt][0pt]{%
            \rule{\arrayrulewidth}{\dimexpr\aboverulesep+\cmidrulewidth+\belowrulesep+0.2pt\relax}%
        }}%
        \vrule width\arrayrulewidth
    }}
    \newcommand{\forecastRows}{%
        \toprule
        \multirow{2}{*}{\forecastHead{\textbf{Method}}} &
        \multicolumn{3}{c!{\forecastHeaderDivider}}{\hspace*{\gapMethodHand}\forecastHead{\textbf{Overall}}\hspace*{\gapPenDivider}} &
        \forecastHead{\mbox{\textbf{Establish Contact}}} &
        \forecastHead{\textbf{Move}} &
        \forecastHead{\textbf{Manipulate}} &
        \forecastHead{\textbf{Put Down}} \\
        \cmidrule(l{\dimexpr\gapMethodHand+\forecastRuleInset\relax}r{\dimexpr\gapPenDivider+\forecastRuleInset+\arrayrulewidth\relax}){2-4}
        \cmidrule(l{\dimexpr\gapDividerReach+\forecastRuleInset\relax}r{\forecastRuleInset}){5-5}
        \cmidrule(l{\dimexpr\gapReachGoal+\forecastRuleInset\relax}r{\forecastRuleInset}){6-6}
        \cmidrule(l{\dimexpr\gapGoalManipulate+\forecastRuleInset\relax}r{\forecastRuleInset}){7-7}
        \cmidrule(l{\dimexpr\gapManipulatePlace+\forecastRuleInset\relax}r{\forecastRuleInset}){8-8}
        & \forecastMetricHead{H-ADE\forecastPairSep{}FDE}{\downarrow}
        & \forecastMetricHead{O-ADE\forecastPairSep{}FDE}{\downarrow}
        & \forecastMetricHead{HO-Pen}{\downarrow}
        & \forecastMetricHead{Reach-SR}{\uparrow}
        & \forecastMetricHead{Goal-DE}{\downarrow}
        & \forecastMetricHead{RA-MPJPE}{\downarrow}
        & \forecastMetricHead{Place-Hit}{\uparrow} \\
        \specialrule{\lightrulewidth}{0.45ex}{0.45ex}
        \forecastMethod{USST}{Bao2023USST} & 5.90\forecastPairSep{}7.97 & \forecastNA & \forecastNA & \forecastNA & \forecastNA & \forecastNA & \forecastNA \\
        \forecastMethod{EgoH4$^{*}$}{Hatano2025EgoH4} & 10.19\forecastPairSep{}11.60 & \forecastNA & \forecastNA & 38.82 & \forecastNA & 3.92 & \forecastNA \\
        \forecastMethod{ObjectForesight}{Soraki2026ObjectForesight} & \forecastNA & 0.90\forecastPairSep{}1.09 & \forecastNA & \forecastNA & 3.79 & \forecastNA & 42.29 \\
        \forecastMethod{MDM$^{\dagger}$}{Tevet2023MDM} & 13.13\forecastPairSep{}14.62 & 22.33\forecastPairSep{}22.95 & 2.44 & 27.80 & 11.25 & 3.59 & 0.44 \\
        \forecastMethod{CAHMP$^{\dagger}$}{Corona2020ContextAware} & 6.34\forecastPairSep{}9.80 & 2.17\forecastPairSep{}3.48 & 10.20 & 47.70 & 7.51 & 1.81 & 14.10 \\
        \forecastMethod{Transformer$^{\ddagger}$}{Vaswani2017Transformer} & 6.43\forecastPairSep{}8.50 & 2.36\forecastPairSep{}2.91 & 13.47 & 38.98 & 7.54 & 1.51 & 11.45 \\
        \forecastMethod{GRU Predictor$^{\ddagger}$}{Cho2014GRU} & 6.73\forecastPairSep{}8.95 & 2.41\forecastPairSep{}3.08 & 14.90 & 40.30 & 7.91 & 1.61 & 13.66 \\
        \midrule
        \rowcolor{ForecastOursRow}
        \forecastOursCell{\textbf{Gaze2HOI (ours)}} &
        \forecastOursCell{\textbf{4.16}\forecastPairSep{}\textbf{5.49}} &
        \forecastOursCell{\textbf{0.72}\forecastPairSep{}\textbf{0.87}} &
        \forecastOursCell{\textbf{1.70}} &
        \forecastOursCell{\textbf{92.43}} &
        \forecastOursCell{\textbf{2.29}} &
        \forecastOursCell{\textbf{1.33}} &
        \forecastOursCell{\textbf{44.93}} \\
        \bottomrule
    }
    \newcommand{\forecastTabular}{%
    \begin{tabular}{l
        >{\hspace*{\gapMethodHand}}c
        >{\hspace*{\gapHandObject}}c
        >{\hspace*{\gapObjectPen}}c<{\hspace*{\gapPenDivider}}
        !{\forecastDivider}
        >{\hspace*{\gapDividerReach}}c
        >{\hspace*{\gapReachGoal}}c
        >{\hspace*{\gapGoalManipulate}}c
        >{\hspace*{\gapManipulatePlace}}c}
        \forecastRows
    \end{tabular}%
    }
    \ifdefined\forecastMeasureBox
    \else
        \newsavebox{\forecastMeasureBox}
    \fi
    
    \sbox{\forecastMeasureBox}{\forecastTabular}
    
    \ifdim\wd\forecastMeasureBox>\linewidth
        \resizebox{\linewidth}{!}{\usebox{\forecastMeasureBox}}%
    \else
        \usebox{\forecastMeasureBox}%
    \fi
    \par\vspace{6pt}
    \begin{minipage}{\linewidth}
        \linespread{1}\fontsize{9pt}{12pt}\selectfont
        $^{*}$Task adaptation and MANO fitting for Reach-SR.
        $^{\dagger}$Adapted for joint hand-object forecasting.
        $^{\ddagger}$Baselines built on the cited architectures.
        \forecastNA: unsupported. \textbf{Bold} denotes the best result. Best of five is selected per test window and metric (min for $\downarrow$, max for $\uparrow$).
        \par
    \end{minipage}
    \endgroup
\end{table*}

\section{Experiments}
This section systematically evaluates Gaze2HOI on EgoAIM through quantitative and qualitative analyses.
These analyses evaluate end-to-end interaction forecasting across phases, analyze the semantic disambiguation, and visualize the geometric disambiguation.
Further experiments examine the effect of intention conditioning on motion forecasting, while ablation studies quantify the contributions of key components.

\subsection{Implementation Details}
The EgoAIM dataset is partitioned into 5,132 training, 976 validation, and 1,321 test windows, ensuring strict temporal separation with no frame overlap between the training and test sets.
Each window comprises 30 observed and 30 future frames at 30 fps, corresponding to one second of observation and one second of forecasting (Fig.~\ref{fig3}).
All methods are evaluated on the same test set, with overall results computed on the full set and phase-specific results on corresponding subsets.

Training is conducted on four NVIDIA RTX A4500 GPUs (with 20GB VRAM each).
Gaze2HOI employs Qwen3-VL-8B-Instruct for intention inference and an 8-block diffusion transformer with a hidden dimension of 384 for coordinated motion forecasting.
Motion generation utilizes 50-step DDIM sampling with a classifier-free guidance scale of 1.5.
For stochastic methods, metric-wise best-of-five results are reported using fixed sampling seeds, whereas deterministic methods retain a single prediction.

\subsection{Evaluation of Egocentric HOI Forecasting}
\subsubsection{Baselines and Metrics}
We benchmark Gaze2HOI against USST~\citep{Bao2023USST} and EgoH4~\citep{Hatano2025EgoH4} for hand forecasting, ObjectForesight~\citep{Soraki2026ObjectForesight} for object forecasting, alongside MDM~\citep{Tevet2023MDM} and CAHMP~\citep{Corona2020ContextAware} adapted for joint hand-object forecasting.
Two additional baselines, GRU Predictor~\citep{Cho2014GRU} and Transformer~\citep{Vaswani2017Transformer}, directly predict joint hand-object states from observed motion and gaze.
All methods are configured for the identical 30-to-30-frame forecasting task based on their supported modalities. 
Omitted metrics due to incompatible outputs are denoted by an em dash.

Overall performance is quantified via average and final displacement errors for hand wrists (H-ADE/FDE) and object translations (O-ADE/FDE), alongside hand-object penetration (HO-Pen).
ADE averages position errors over valid future frames, whereas FDE evaluates the final valid frame.
Object errors are averaged equally over all distinct scene objects with valid observed poses; objects outside Gaze2HOI's predicted bindings retain their last observed poses.
HO-Pen calculates the percentage of valid future frames where either hand penetrates a scene object by a margin exceeding 1.5 cm.
Distance errors are consistently reported in centimeters, while all rates are expressed as percentages.

\begin{figure*}[!t]
    \centering
    \includegraphics[width=\textwidth]{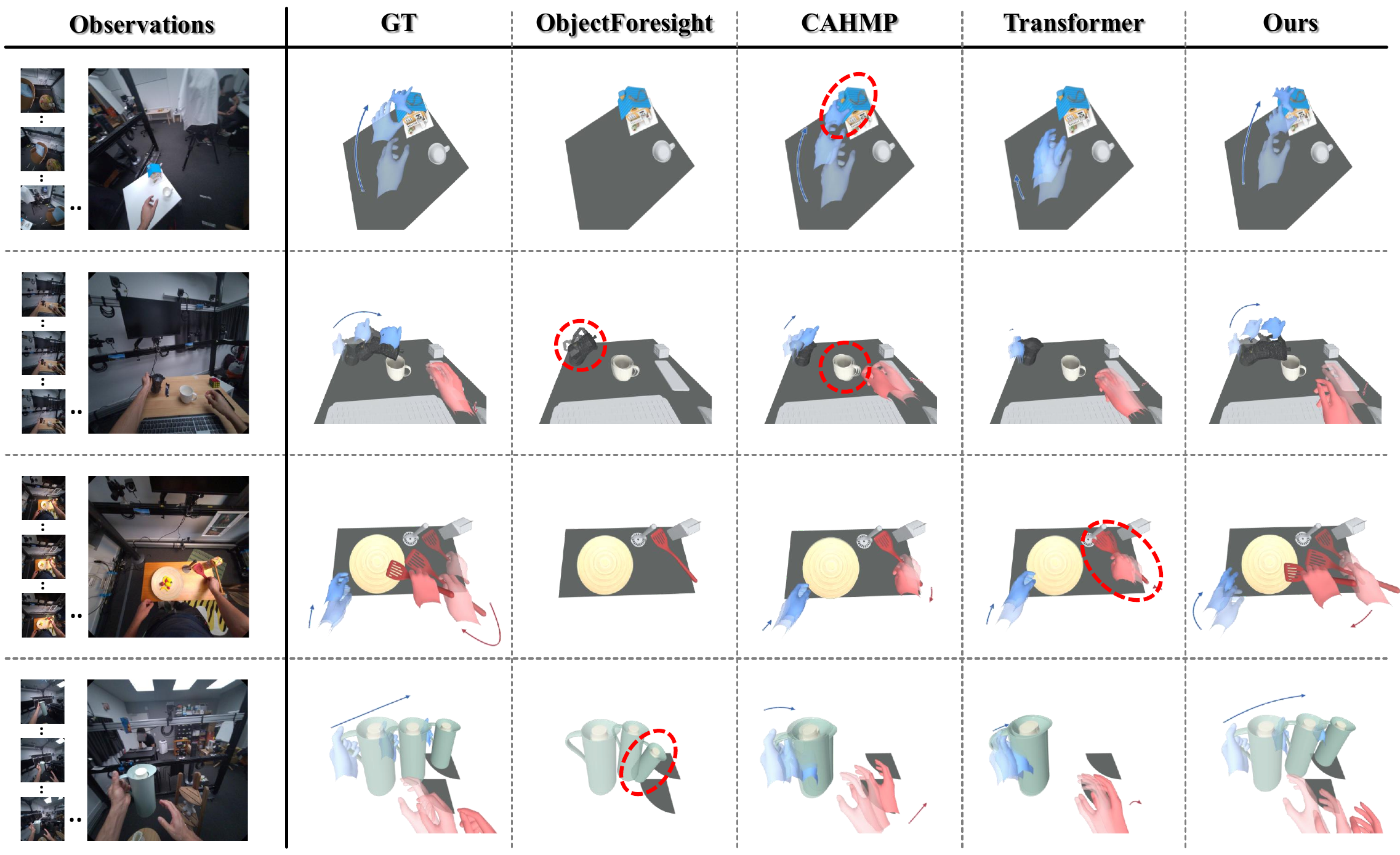}
    \caption{\textbf{\emph{Qualitative comparison of hand-object interaction forecasting on EgoAIM.}}
    From top to bottom, the rows correspond to \emph{Establish Contact}, \emph{Move}, \emph{Manipulate}, and \emph{Put Down}.
    Observed inputs and ground-truth (GT) futures are shown alongside predictions from representative baselines and Gaze2HOI.}
    \label{fig5}
\end{figure*}

Each interaction phase is further assessed using a dedicated metric.
For \emph{Establish Contact}, Reach-SR is the percentage of reaches for which any predicted hand vertex comes within 2 cm of the ground-truth task-object surface.
For \emph{Move}, Goal-DE quantifies the absolute error of the terminal object-to-target distance, with the target fixed at its last observed position.
For \emph{Manipulate}, RA-MPJPE computes the mean 3D joint error after wrist translation alignment.
For \emph{Put Down}, Place-Hit measures the percentage of task objects whose predicted endpoint lies within 5 cm of the ground-truth endpoint.
These metrics are computed on the corresponding phase subsets using annotated active hands or task objects.

\subsubsection{Quantitative Results}
As detailed in Table~\ref{tab2}, Gaze2HOI achieves the best reported results across all metrics for joint hand-object forecasting.
Relative to USST, it reduces H-ADE and H-FDE by 29.5\% and 31.1\%, respectively, while reducing O-ADE and O-FDE by 20.0\% and 20.2\% relative to ObjectForesight.
HO-Pen decreases to 1.70\%, compared with 2.44\% for MDM.
Despite its comparatively low penetration rate, the MDM adaptation exhibits substantially larger trajectory errors, underscoring the necessity of jointly evaluating geometric accuracy and physical plausibility.

The phase-specific results show a particularly large improvement in \emph{Establish Contact}, where Reach-SR reaches 92.43\%, exceeding CAHMP by 44.73 percentage points.
For \emph{Move} and \emph{Manipulate}, Gaze2HOI achieves a Goal-DE of 2.29 cm and an RA-MPJPE of 1.33 cm, reducing the errors of ObjectForesight and Transformer by 39.6\% and 11.9\%, respectively.
The improvement in \emph{Put Down} is more modest: Place-Hit increases from ObjectForesight's 42.29\% to 44.93\%, yielding a gain of 2.64 percentage points.
The remaining endpoint errors identify precise placement as an area for further improvement.

\begin{figure*}[!t]
    \centering
    \includegraphics[width=\textwidth]{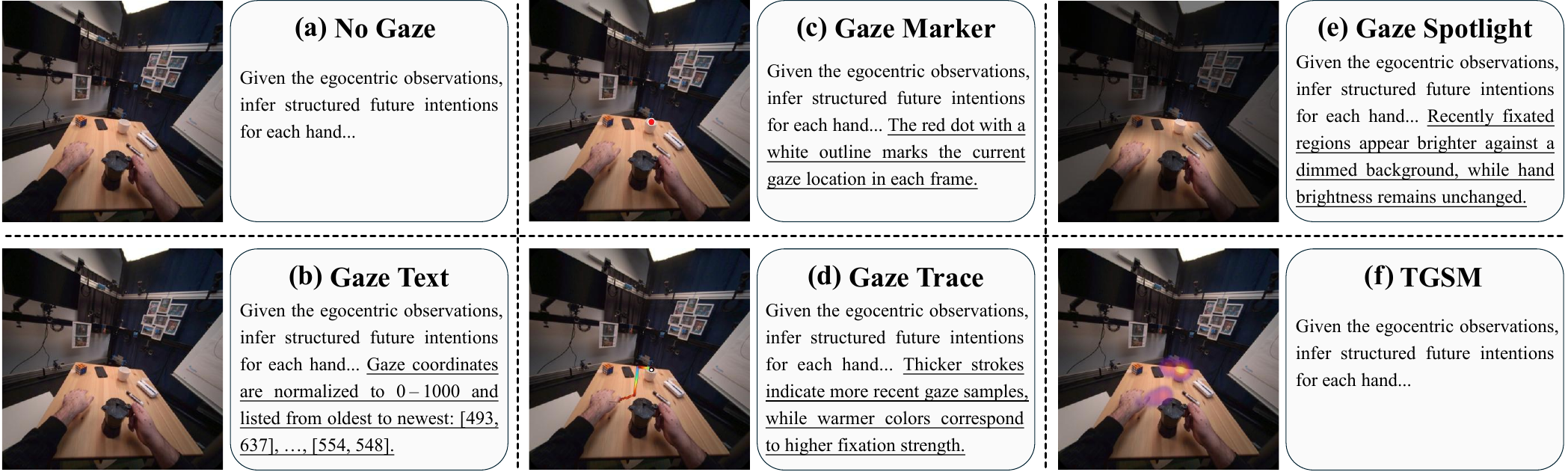}
    \caption{\textbf{\emph{Illustration of gaze injection strategies.}}
    Each panel shows the image and corresponding prompt for the same egocentric observation.
    The TGSM overlay is for visualization only; gaze cues are injected through visual feature modulation.}
    \label{fig6}
    \vspace{-2 mm}
\end{figure*}

\subsubsection{Qualitative Results}
The qualitative comparisons in Fig.~\ref{fig5} detail the representative failure modes of existing methods.
Several baselines capture coarse motion trends but exhibit errors in contact configuration, motion extent, or object orientation.
The moka pot example illustrates that plausible object poses alone do not ensure goal-directed interaction: ObjectForesight fails to position it above the cup.
Transformer exhibits a related shortcoming, predicting insufficient spatula movement toward the plate.
Beyond reaching the target region, successful hand-object interaction also requires appropriate contact configurations and object orientations.
CAHMP approaches the toy house with inaccurate finger articulation near the contact surface, while ObjectForesight transports the pitcher with excessive tilt.
The highlighted cup reveals another issue: CAHMP predicts drift despite the object remaining nearly stationary in the ground truth.
These examples underscore the importance of jointly capturing goal-directed motion, stable hand-object relationships, and scene consistency.
Gaze2HOI overcomes these limitations, generating coherent hand-object interactions with coordinated articulation and object motion.

\begin{table}[!t]
    \centering
    \caption{Comparison of gaze injection strategies.}
    \label{tab3}
    \begingroup
    \linespread{1}\fontsize{8.2pt}{11pt}\selectfont
    \setlength{\tabcolsep}{2pt}
    \renewcommand{\arraystretch}{1.08}
    \setlength{\aboverulesep}{0.2ex}
    \setlength{\belowrulesep}{0.2ex}
    \definecolor{GazeOursRow}{RGB}{242,242,242}

    \newcommand{\gazeHead}[1]{%
        \begingroup
        \fontsize{7pt}{9pt}\selectfont
        \renewcommand{\arraystretch}{1}
        \begin{tabular}[c]{@{}c@{}}#1\end{tabular}%
        \endgroup
    }

    \begin{tabularx}{\linewidth}{
        @{}l
        *{3}{>{\centering\arraybackslash}X}
        @{}
    }
        \toprule
        \gazeHead{\textbf{Method}} &
        \gazeHead{Active-hand\\EM (\%)\,$\uparrow$} &
        \gazeHead{Median Point\\Error (px)\,$\downarrow$} &
        \gazeHead{Sample EM\\@100px (\%)\,$\uparrow$} \\
        \specialrule{\lightrulewidth}{0.45ex}{0.45ex}
        No Gaze        & 81.76 & 51.2 & 72.67 \\
        Gaze Text      & 83.95 & 38.0 & 77.44 \\
        Gaze Marker    & 86.20 & 39.2 & 80.85 \\
        Gaze Trace     & 85.05 & 40.2 & 78.35 \\
        Gaze Spotlight & 83.22 & 50.7 & 76.00 \\
        \midrule
        \rowcolor{GazeOursRow}
        \textbf{TGSM (ours)} &
        \textbf{87.84} &
        \textbf{35.2} &
        \textbf{82.29} \\
        \bottomrule
    \end{tabularx}
    \endgroup
    \vspace{-4 mm}
\end{table}

\subsection{Analysis of Semantic Disambiguation}
\subsubsection{Evaluation Protocol}
To quantify semantic disambiguation, we evaluate the accuracy of structured intention inference using three complementary metrics: Active-hand Exact Match (EM), Median Point Error, and Sample EM@100px.
Active-hand EM measures the percentage of annotated active hands for which the complete set of categorical intention attributes is correctly predicted.
Median Point Error reports the median Euclidean localization error in the original $1408\times1408$ image space, considering only cases with both an annotated and a predicted destination.
Sample EM@100px measures the percentage of samples with correct categorical predictions for both hands and destination errors within 100 pixels.
This sample-level metric also requires correct predictions of inactive hands and absent destination points.
Together, these metrics assess semantic accuracy, spatial precision, and complete intention prediction.

\begin{table}[!t]
    \centering
    \caption{Intention inference performance across VLMs.}
    \label{tab4}
    \begingroup
    \linespread{1}\fontsize{8.2pt}{11pt}\selectfont
    \setlength{\tabcolsep}{2pt}
    \renewcommand{\arraystretch}{1.08}
    \setlength{\aboverulesep}{0.2ex}
    \setlength{\belowrulesep}{0.2ex}

    \definecolor{IntentionOursRow}{RGB}{242,242,242}

    \newcommand{\intentHead}[1]{%
        \begingroup
        \fontsize{8pt}{9pt}\selectfont
        \renewcommand{\arraystretch}{1}
        \begin{tabular}[c]{@{}c@{}}#1\end{tabular}%
        \endgroup
    }

    \begin{tabularx}{\linewidth}{
        l
        *{2}{>{\centering\arraybackslash}X}
    }
        \toprule
        \intentHead{\textbf{Model}} &
        \intentHead{Gaze Injection} &
        \intentHead{Sample EM\\@100px (\%)\,$\uparrow$} \\
        \specialrule{\lightrulewidth}{0.45ex}{0.45ex}

        GPT-5.4        & Gaze Marker & 39.21 \\
        Gemini 3.1 Pro & Gaze Marker & 25.66 \\

        \noalign{%
            \vskip 0.4ex
            \hbox to \linewidth{%
                \color{black!40}%
                \cleaders
                \hbox{\rule{1.5pt}{0.35pt}\hskip 2pt}%
                \hfill
            }%
            \vskip 0.4ex
        }

        Qwen2-VL-2B & TGSM & 68.81 \\
        Qwen2-VL-7B & TGSM & 77.97 \\
        Qwen3-VL-2B & TGSM & 69.04 \\
        Qwen3-VL-4B & TGSM & 73.81 \\

        \midrule
        \rowcolor{IntentionOursRow}
        \textbf{Qwen3-VL-8B (ours)} &
        TGSM &
        \textbf{82.29} \\
        \bottomrule
    \end{tabularx}
    \endgroup
    \vspace{-6 mm}
\end{table}

\subsubsection{Gaze Injection Strategies}
We compare six gaze injection strategies using the same Qwen3-VL-8B backbone, data splits, and training schedule.
As illustrated in Fig.~\ref{fig6}, \emph{No Gaze} processes unmodified observations, whereas \emph{Gaze Text} provides gaze coordinates and \emph{Gaze Marker} overlays the current gaze location.
\emph{Gaze Trace} encodes temporal recency and fixation strength through trajectory width and color, respectively.
\emph{Gaze Spotlight} uses these cues to modulate spatial brightness.
These four gaze baselines operate at the input level.
In contrast, TGSM incorporates temporal gaze saliency via visual feature modulation; its overlay in Fig.~\ref{fig6} is provided solely to visualize the gaze cues used for conditioning.

As shown in Table~\ref{tab3}, TGSM achieves the best performance across all three metrics, reaching 87.84\% Active-hand EM and 82.29\% Sample EM@100px, alongside a median point error of 35.2 pixels.
Compared with \emph{No Gaze}, TGSM improves Sample EM@100px by 9.62 percentage points.
\emph{Gaze Marker} achieves the highest Sample EM@100px among the input-level gaze injection strategies, but remains 1.44 percentage points below TGSM.
The suboptimal scores of \emph{Gaze Trace} and \emph{Gaze Spotlight} relative to \emph{Gaze Marker} suggest that explicitly rendering temporal gaze information at the input level does not necessarily lead to more accurate intention inference.
These results validate TGSM as an effective means of incorporating temporal gaze cues while preserving the original visual observations.

\subsubsection{Comparison across VLMs}
We compare two commercial VLMs with five Qwen configurations using Sample EM@100px on the same test set.
GPT-5.4 and Gemini 3.1 Pro receive Gaze Marker inputs without task-specific fine-tuning, while the Qwen models are fine-tuned with TGSM.
All models are instructed to produce the same structured output format.

As reported in Table~\ref{tab4}, GPT-5.4 and Gemini 3.1 Pro achieve 39.21\% and 25.66\%, respectively, while the task-adapted Qwen models attain substantially higher accuracy.
Within the Qwen3-VL family, performance improves consistently with model size, increasing from 69.04\% at 2B to 73.81\% at 4B and 82.29\% at 8B.
Qwen2-VL-7B achieves 77.97\%, outperforming the smaller Qwen3-VL variants but trailing Qwen3-VL-8B by 4.32 percentage points.
These results support the selection of Qwen3-VL-8B as the intention inference backbone of Gaze2HOI.

\par
\vspace{5 mm}

\begin{figure}[!t]
    \centering
    \includegraphics[width=\columnwidth]{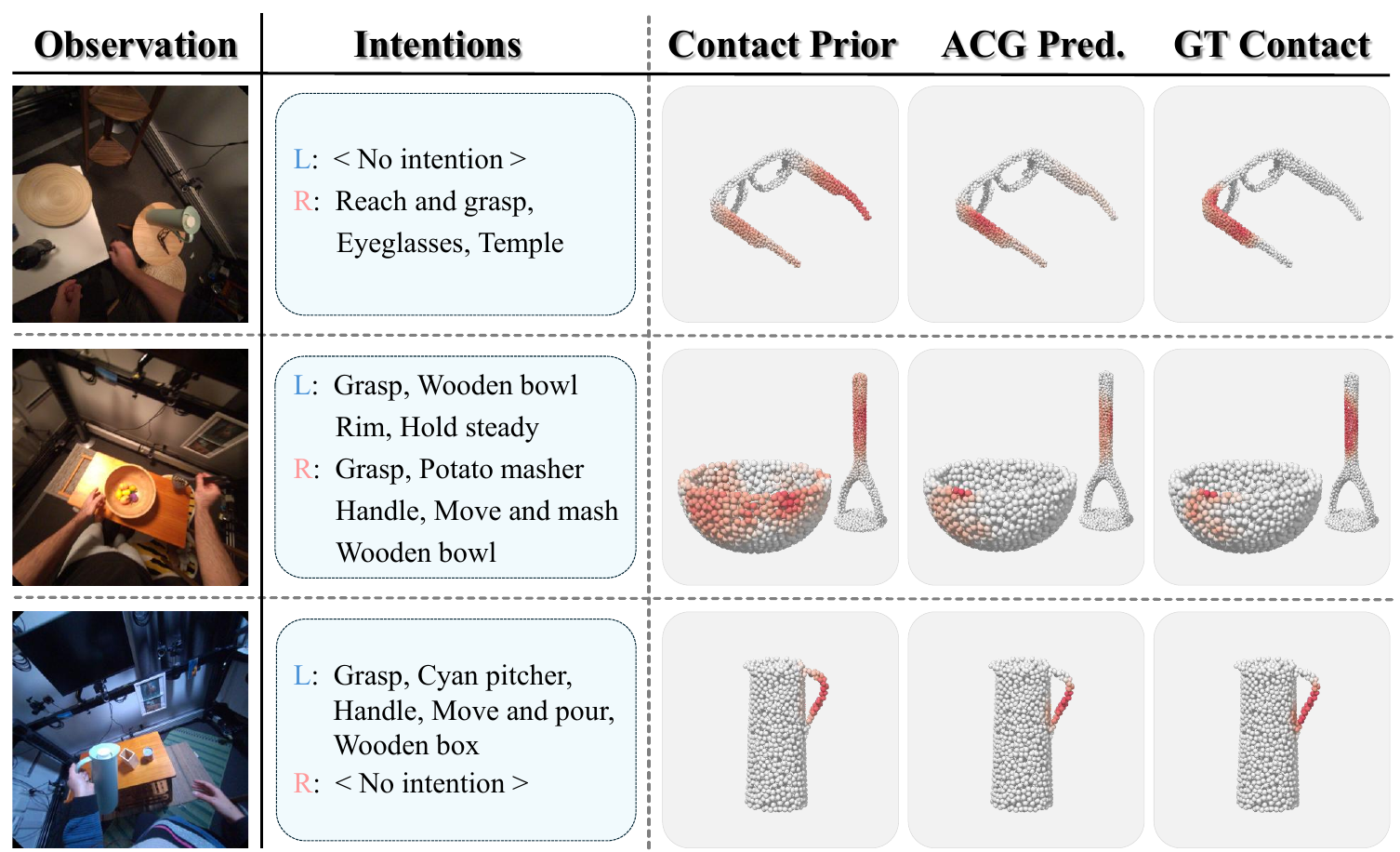}
    \caption{\textbf{\emph{Visualization of geometric disambiguation via ACG.}}
    ACG predicts future contact regions using contact priors and observed interactions. 
    Warmer colors indicate higher relative contact intensity within each map.
    }
    \label{fig7}
    \vspace{4 mm}
\end{figure}

\subsection{Visualization of Geometric Disambiguation}
We qualitatively examine how ACG uses observed interactions to refine contact priors, thereby achieving geometric disambiguation in future contact localization.
As illustrated in Fig.~\ref{fig7}, the contact priors identify plausible contact regions associated with the intended object and part, while ACG further resolves their spatial extent using the observed hand-object states and hand trajectories.
For the eyeglasses, the predicted contact probability becomes concentrated on one temple.
For the bowl, masher, and pitcher, ACG narrows the broader rim and handle priors to localized regions relevant to the ongoing interaction.
These examples illustrate how observed interaction cues complement contact priors, enabling more precise contact localization and providing finer geometric guidance for subsequent motion forecasting.

\par
\vspace{5 mm}

\subsection{Effectiveness of Intention in Motion Forecasting}
To assess the value of inferred intentions for downstream motion forecasting, we compare three conditions: no explicit intentions or destinations (\emph{None}), predicted intentions and destinations (\emph{Predicted}), and ground-truth annotations (\emph{Oracle}).
The motion model and ACG weights remain fixed across all settings.
For \emph{None}, ACG provides contact guidance from observed interactions without using explicit intention or destination information.
All settings follow the same metric-wise best-of-five evaluation protocol.

\begin{table}[!htbp]
    \centering
    \caption{Motion forecasting across intention conditions.}
    \label{tab5}
    \begingroup
    \footnotesize
    \renewcommand{\arraystretch}{1.35}
    \setlength{\aboverulesep}{0.2ex}
    \setlength{\belowrulesep}{0.2ex}
    \definecolor{IntentionConditionRow}{RGB}{242,242,242}

    \newcommand{\conditionFirstGap}{4pt}
    \newcommand{\conditionHead}[1]{\mbox{#1}}

    \newcommand{\conditionDelta}[1]{%
        \raisebox{-0.8pt}{%
            \fontsize{5.5pt}{6.5pt}\selectfont
            \color{black!55}#1%
        }%
    }

    \ifdefined\conditionAdeDeltaBox\else
        \newsavebox{\conditionAdeDeltaBox}
    \fi
    \ifdefined\conditionRateDeltaBox\else
        \newsavebox{\conditionRateDeltaBox}
    \fi
    \ifdefined\conditionMeasureBox\else
        \newsavebox{\conditionMeasureBox}
    \fi
    \ifdefined\conditionHeaderBox\else
        \newsavebox{\conditionHeaderBox}
    \fi
    \ifdefined\conditionCellBox\else
        \newsavebox{\conditionCellBox}
    \fi

    \sbox{\conditionAdeDeltaBox}{%
        \conditionDelta{$-47.6\%$}%
    }
    \sbox{\conditionRateDeltaBox}{%
        \conditionDelta{$+76.7\,\mathrm{pp}$}%
    }

    \newcommand{\conditionValue}[3][\conditionAdeDeltaBox]{%
        #2\hspace{0.3pt}%
        \makebox[\wd#1][l]{%
            \conditionDelta{#3}%
        }%
    }

    \newcommand{\conditionRate}[2]{%
        \conditionValue[\conditionRateDeltaBox]{#1}{#2}%
    }

    \newcommand{\conditionWidest}[2]{%
        \sbox{\conditionHeaderBox}{#1}%
        \sbox{\conditionCellBox}{#2}%
        \ifdim\wd\conditionHeaderBox>\wd\conditionCellBox
            \usebox{\conditionHeaderBox}%
        \else
            \usebox{\conditionCellBox}%
        \fi
    }

    \sbox{\conditionMeasureBox}{%
        \conditionWidest{Intention}{Predicted}%
        \hspace*{\conditionFirstGap}%
        \conditionWidest
            {\conditionHead{H-ADE\,$\downarrow$}}
            {\conditionValue{6.49}{}}%
        \conditionWidest
            {\conditionHead{O-ADE\,$\downarrow$}}
            {\conditionValue{1.05}{}}%
        \conditionWidest
            {\conditionHead{Reach-SR\,$\uparrow$}}
            {\conditionRate{96.22}{}}%
        \conditionWidest
            {\conditionHead{Place-Hit\,$\uparrow$}}
            {\conditionRate{95.59}{}}%
    }

    \setlength{\tabcolsep}{0pt}
    \ifdim\wd\conditionMeasureBox>\columnwidth
        \PackageWarning{Gaze2HOI}{%
            Intention table exceeds column width%
        }%
    \else
        \setlength{\tabcolsep}{%
            \dimexpr(\columnwidth-\wd\conditionMeasureBox)/10\relax
        }
    \fi

    \begin{tabular}{l<{\hspace*{\conditionFirstGap}}llll}
        \toprule
        Intention &
        \conditionHead{H-ADE\,$\downarrow$} &
        \conditionHead{O-ADE\,$\downarrow$} &
        \conditionHead{Reach-SR\,$\uparrow$} &
        \conditionHead{Place-Hit\,$\uparrow$} \\
        \specialrule{\lightrulewidth}{0.45ex}{0.45ex}

        None &
        \conditionValue{6.49}{} &
        \conditionValue{1.05}{} &
        \conditionRate{56.91}{} &
        \conditionRate{18.94}{} \\

        \rowcolor{IntentionConditionRow}
        Predicted &
        \conditionValue{4.16}{$-35.9\%$} &
        \conditionValue{0.72}{$-31.4\%$} &
        \conditionRate{92.43}{$+35.5\,\mathrm{pp}$} &
        \conditionRate{44.93}{$+26.0\,\mathrm{pp}$} \\

        Oracle &
        \conditionValue{3.65}{$-43.8\%$} &
        \conditionValue{0.55}{$-47.6\%$} &
        \conditionRate{96.22}{$+39.3\,\mathrm{pp}$} &
        \conditionRate{95.59}{$+76.7\,\mathrm{pp}$} \\
        \bottomrule
    \end{tabular}

    \par\vspace{6pt}
    \begin{minipage}{\columnwidth}
        \linespread{1}\fontsize{7pt}{10pt}\selectfont
        Changes vs.\ None: relative \% for ADE;
        percentage points (pp) for success rates.
        \par
    \end{minipage}
    \endgroup
    \vspace{-4 mm}
\end{table}

\begin{figure*}[!t]
    \centering
    \includegraphics[width=0.98\textwidth]{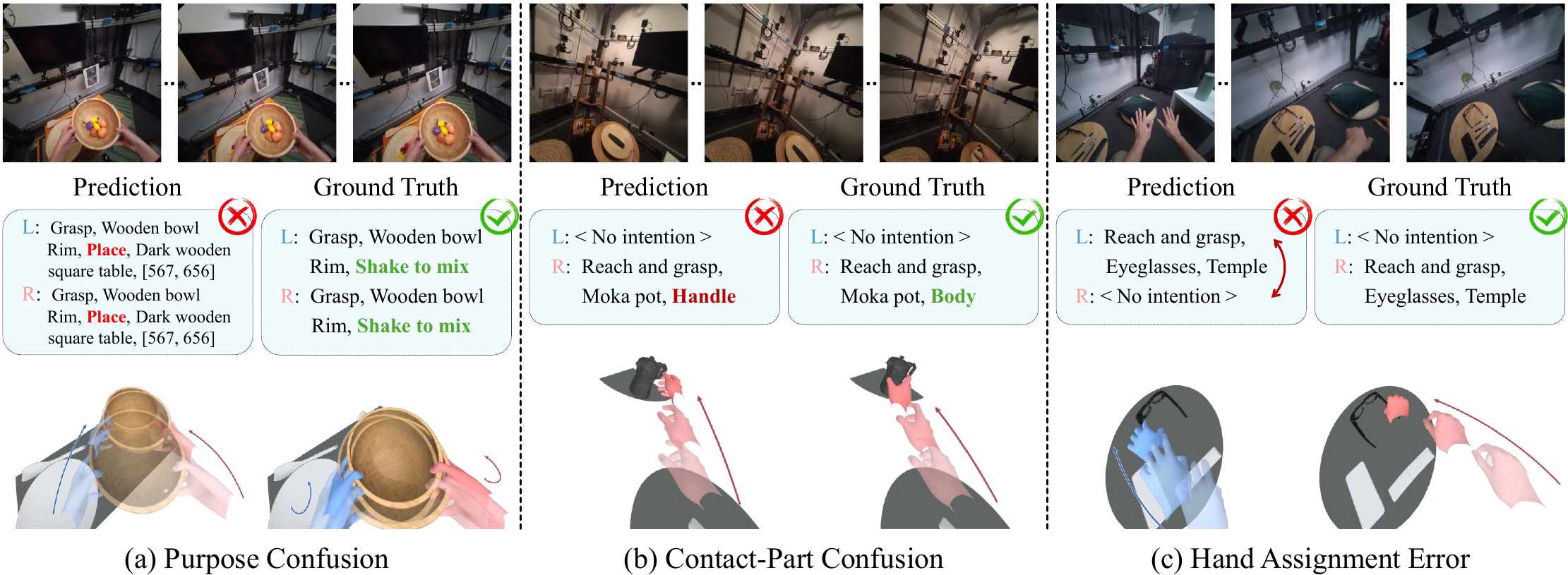}
    \caption{\textbf{\emph{Failure cases of Gaze2HOI.}}
    Examples of (a) purpose confusion, (b) contact-part confusion, and (c) hand assignment errors, with predicted intentions and the corresponding future interactions shown alongside ground truth.}
    \label{fig8}
\end{figure*}

As shown in Table~\ref{tab5}, predicted intentions substantially improve both trajectory accuracy and interaction success over intention-free conditioning.
Relative to \emph{None}, \emph{Predicted} reduces H-ADE and O-ADE by 35.9\% and 31.4\%, respectively, demonstrating benefits for both hand and object forecasting.
These trajectory improvements are accompanied by a 35.5-percentage-point increase in Reach-SR, which reaches 92.43\% and approaches the oracle result of 96.22\%.
Placement also benefits considerably, with Place-Hit increasing from 18.94\% to 44.93\%; however, the remaining gap to the oracle result of 95.59\% indicates that precise placement remains challenging under predicted intention conditioning.

\subsection{Ablation Studies}
We assess the contributions of gaze information, TGSM, and geometric conditioning to end-to-end hand-object motion forecasting through the ablations in Table~\ref{tab6}.
The full model achieves the best results across all four reported metrics.
Removing gaze causes the largest degradation, reducing Reach-SR and Place-Hit by 11.18 and 18.06 percentage points, respectively.
Replacing TGSM with Gaze Marker recovers much of the contact-establishment performance, reaching 91.61\% Reach-SR, but yields a Place-Hit of only 33.48\%, compared with 44.93\% for the full model.
These results demonstrate that the benefits of TGSM extend beyond intention inference to the accuracy of downstream interaction outcomes.

\begin{table}[!t]
    \centering
    \caption{End-to-end ablation study of Gaze2HOI.}
    \label{tab6}
    \begingroup
    \footnotesize
    \setlength{\tabcolsep}{2.5pt}
    \renewcommand{\arraystretch}{1.12}
    \setlength{\aboverulesep}{0.2ex}
    \setlength{\belowrulesep}{0.2ex}
    \definecolor{AblationOursRow}{RGB}{242,242,242}

    \begin{tabularx}{\columnwidth}{
        >{\raggedright\arraybackslash}X
        c c c c
    }
        \toprule
        Setting &
        H-ADE\,$\downarrow$ &
        O-ADE\,$\downarrow$ &
        Reach-SR\,$\uparrow$ &
        Place-Hit\,$\uparrow$ \\
        \specialrule{\lightrulewidth}{0.45ex}{0.45ex}

        w/o TGSM & 4.32 & 0.79 & 91.61 & 33.48 \\
        w/o Gaze & 4.65 & 0.87 & 81.25 & 26.87 \\
        w/o ACG  & 4.35 & 0.78 & 88.82 & 33.48 \\

        \specialrule{\lightrulewidth}{0.15ex}{0.2ex}
        \rowcolor{AblationOursRow}
        \textbf{Full Model} &
        \textbf{4.16} &
        \textbf{0.72} &
        \textbf{92.43} &
        \textbf{44.93} \\
        \bottomrule
    \end{tabularx}

    \par\vspace{6pt}
    \begin{minipage}{\columnwidth}
        \linespread{1}\fontsize{7pt}{10pt}\selectfont
        \emph{w/o TGSM} replaces TGSM with the input-level Gaze Marker strategy.
        \emph{w/o Gaze} excludes gaze information from intention inference.
        \emph{w/o ACG} removes all explicit geometric conditioning, including ACG fields and goal encodings.
        \par
    \end{minipage}
    \endgroup
    \vspace{-5 mm}
\end{table}

Removing explicit geometric conditioning, including ACG fields and goal encodings, also degrades forecasting despite retaining structured intentions and observed motion.
H-ADE and O-ADE increase to 4.35 and 0.78 cm, respectively, while Reach-SR falls to 88.82\% and Place-Hit to 33.48\%.
The substantial reduction in placement success, alongside the comparatively modest changes in average displacement error, highlights the importance of geometric and goal information for completing precise interactions.
These conditions complement semantic intentions by specifying spatial constraints that high-level intention labels alone leave underdetermined.

\section{Discussion}
Our results highlight the role of structured intentions in guiding hand-object interaction forecasting.
The performance gains over the no-intention setting (Table~\ref{tab5}) demonstrate that inferred task semantics provide effective guidance even when intention predictions are imperfect.
ACG complements this information by refining semantic contact priors using observed interaction cues, as visualized in Fig.~\ref{fig7}.
Together, these findings support integrating semantic reasoning with geometric grounding, with performance under predicted intentions approaching that achieved under oracle conditioning in specific interaction phases.

Interpreting the failure cases in Fig.~\ref{fig8} requires distinguishing true goal misalignment from potentially valid interaction variations.
In Fig.~\ref{fig8}(a), predicting \emph{placing} instead of \emph{shaking} fundamentally alters the intended goal.
In Fig.~\ref{fig8}(b,c), however, the action and task object are correctly inferred, while the predicted contact part or active hand differs from the annotation.
These alternative choices may still serve as viable strategies for the same high-level task.
Even for a human observer, the available visual history may not uniquely determine the contact region or executing hand before the interaction unfolds.
Consequently, although these predictions are penalized by annotation-based metrics, disagreement with a single recorded future does not necessarily imply failure to capture the underlying interaction goal.

Precise placement remains a substantial challenge.
Although predicted intentions improve Place-Hit from 18.94\% to 44.93\%, the 95.59\% achieved under oracle conditioning reveals a substantial performance gap.
This gap suggests that precise placement is highly sensitive to inaccuracies in the inferred intention conditions.
Future work could model uncertainty in predicted intention attributes and jointly refine spatial goals with contact and support constraints to improve placement accuracy and physical consistency.

\par
\vspace{2 mm}

\section{Conclusion}
In this paper, we presented Gaze2HOI, a hierarchical framework that integrates gaze-aware intention inference, anticipatory contact grounding, and coordinated motion generation for egocentric hand-object interaction forecasting.
TGSM incorporates temporal gaze cues into visual representations to infer structured per-hand intentions, while ACG combines contact priors with observed interactions and object geometry to predict spatially explicit contact, support, and placement fields.
These semantic and geometric conditions guide a diffusion model that jointly forecasts the articulated motion of both hands and 6-DoF object trajectories.
We also introduced EgoAIM, a phase-aware benchmark pairing fine-grained future intentions with continuous 3D hand-object motion, enabling evaluation across four interaction phases.
Experiments on EgoAIM demonstrate improvements over the evaluated baselines in both intention inference and interaction forecasting, supporting the complementary contributions of gaze information and geometric grounding.
End-to-end performance approaches the levels achieved under oracle conditioning in specific phases, while precise placement remains an important direction for future improvement.
By connecting inferred interaction goals with their anticipated spatial realization, this work provides a foundation for advancing proactive human-centric assistance in extended reality and embodied systems.

\section*{Declarations}

\bmhead*{Funding}
This work was partially supported by the National Natural Science Foundation of China (Grant No. 62306309).

\bmhead*{Competing interests}
The authors declare that they have no competing interests.

\bmhead*{Data availability}
EgoAIM is derived from \href{https://facebookresearch.github.io/hot3d/}{HOT3D}. Representative examples are publicly available in the \href{https://huggingface.co/datasets/gaze2hoi/EgoAIM-Demo}{EgoAIM-Demo} repository. Following acceptance of this article, the full EgoAIM annotations and data splits will be available from the corresponding author upon reasonable request. Data use remains subject to the applicable source-dataset licenses.

\bmhead*{Code availability}
Inference code and pretrained checkpoints are publicly available on \href{https://github.com/gaze2hoi/Gaze2HOI}{GitHub} and \href{https://huggingface.co/gaze2hoi/Gaze2HOI}{Hugging Face}, respectively.

\bibliography{sn-bibliography}

@article{Plizzari2024EgocentricSurvey,
  author    = {Plizzari, Chiara and Goletto, Gabriele and Furnari, Antonino and Bansal, Siddhant and Ragusa, Francesco and Farinella, Giovanni Maria and Damen, Dima and Tommasi, Tatiana},
  title     = {An Outlook into the Future of Egocentric Vision},
  journal   = {International Journal of Computer Vision (IJCV)},
  year      = {2024},
  volume    = {132},
  number    = {11},
  pages     = {4880--4936},
  doi       = {10.1007/s11263-024-02095-7}
}

@article{Xu2023DCR,
  author    = {Xu, Xinyu and Li, Yong-Lu and Lu, Cewu},
  title     = {Dynamic Context Removal: A General Training Strategy for Robust Models on Video Action Predictive Tasks},
  journal   = {International Journal of Computer Vision (IJCV)},
  year      = {2023},
  volume    = {131},
  number    = {12},
  pages     = {3272--3288},
  doi       = {10.1007/s11263-023-01850-6}
}

@article{Stergiou2025AboutTime,
  author    = {Stergiou, Alexandros and Poppe, Ronald},
  title     = {About Time: Advances, Challenges, and Outlooks of Action Understanding},
  journal   = {International Journal of Computer Vision (IJCV)},
  year      = {2025},
  volume    = {133},
  number    = {9},
  pages     = {6251--6315},
  doi       = {10.1007/s11263-025-02478-4}
}

@article{Damen2022EPICKITCHENS100,
  author    = {Damen, Dima and Doughty, Hazel and Farinella, Giovanni Maria and Furnari, Antonino and Kazakos, Evangelos and Ma, Jian and Moltisanti, Davide and Munro, Jonathan and Perrett, Toby and Price, Will and Wray, Michael},
  title     = {Rescaling Egocentric Vision: Collection, Pipeline and Challenges for {EPIC-KITCHENS-100}},
  journal   = {International Journal of Computer Vision (IJCV)},
  year      = {2022},
  volume    = {130},
  number    = {1},
  pages     = {33--55},
  doi       = {10.1007/s11263-021-01531-2}
}

@article{Sui2026InteractionMotionSurvey,
  author    = {Sui, Kewei and Ghosh, Anindita and Hwang, Inwoo and Zhou, Bing and Wang, Jian and Guo, Chuan},
  title     = {A Survey on Human Interaction Motion Generation},
  journal   = {International Journal of Computer Vision (IJCV)},
  year      = {2026},
  volume    = {134},
  number    = {3},
  pages     = {113},
  doi       = {10.1007/s11263-025-02582-5}
}

@article{Qiu2026EgoPlanBench2,
  author    = {Qiu, Lu and Chen, Yi and Ge, Yuying and Ge, Yixiao and Shan, Ying and Liu, Xihui},
  title     = {{EgoPlan-Bench2}: A Benchmark for Multimodal Large Language Model Planning in Real-World Scenarios},
  journal   = {International Journal of Computer Vision (IJCV)},
  year      = {2026},
  volume    = {134},
  number    = {5},
  pages     = {222},
  doi       = {10.1007/s11263-026-02826-y}
}

@inproceedings{Driess2023PaLM-E,
  author    = {Driess, Danny and Xia, Fei and Sajjadi, Mehdi S. M. and Lynch, Corey and Chowdhery, Aakanksha and Ichter, Brian and Wahid, Ayzaan and Tompson, Jonathan and Vuong, Quan and Yu, Tianhe and Huang, Wenlong and Chebotar, Yevgen and Sermanet, Pierre and Duckworth, Daniel and Levine, Sergey and Vanhoucke, Vincent and Hausman, Karol and Toussaint, Marc and Greff, Klaus and Zeng, Andy and Mordatch, Igor and Florence, Pete},
  title     = {{PaLM-E}: An Embodied Multimodal Language Model},
  booktitle = {Proceedings of the International Conference on Machine Learning ({ICML})},
  year      = {2023},
  volume    = {202},
  pages     = {8469--8488}
}

@inproceedings{Zheng2022GIMO,
  author    = {Zheng, Yang and Yang, Yanchao and Mo, Kaichun and Li, Jiaman and Yu, Tao and Liu, Yebin and Liu, C. Karen and Guibas, Leonidas J.},
  title     = {{GIMO}: Gaze-Informed Human Motion Prediction in Context},
  booktitle = {Proceedings of the European Conference on Computer Vision ({ECCV})},
  year      = {2022},
  volume    = {13673},
  pages     = {676--694},
  doi       = {10.1007/978-3-031-19778-9_39}
}

@inproceedings{Lee2025SensibleAgent,
  author    = {Lee, Geonsun and Xia, Min and Numan, Nels and Qian, Xun and Li, David and Chen, Yanhe and Kulshrestha, Achin and Chatterjee, Ishan and Zhang, Yinda and Manocha, Dinesh and Kim, David and Du, Ruofei},
  title     = {{Sensible Agent}: A Framework for Unobtrusive Interaction with Proactive {AR} Agents},
  booktitle = {Proceedings of the Annual {ACM} Symposium on User Interface Software and Technology ({UIST})},
  year      = {2025},
  articleno = {119},
  pages     = {1--22},
  doi       = {10.1145/3746059.3747748}
}

@inproceedings{Li2022EgoPAT3D,
  author    = {Li, Yiming and Cao, Ziang and Liang, Andrew and Liang, Benjamin and Chen, Luoyao and Zhao, Hang and Feng, Chen},
  title     = {Egocentric Prediction of Action Target in {3D}},
  booktitle = {Proceedings of the {IEEE/CVF} Conference on Computer Vision and Pattern Recognition ({CVPR})},
  year      = {2022},
  pages     = {20971--20980},
  doi       = {10.1109/CVPR52688.2022.02033}
}

@inproceedings{Liu2024TACO,
  author    = {Liu, Yun and Yang, Haolin and Si, Xu and Liu, Ling and Li, Zipeng and Zhang, Yuxiang and Liu, Yebin and Yi, Li},
  title     = {{TACO}: Benchmarking Generalizable Bimanual {Tool-ACtion-Object} Understanding},
  booktitle = {Proceedings of the Computer Vision and Pattern Recognition ({CVPR})},
  year      = {2024},
  pages     = {21740--21751},
  doi       = {10.1109/CVPR52733.2024.02054}
}

@inproceedings{Liu2022OCT,
  author    = {Liu, Shaowei and Tripathi, Subarna and Majumdar, Somdeb and Wang, Xiaolong},
  title     = {Joint Hand Motion and Interaction Hotspots Prediction from Egocentric Videos},
  booktitle = {Proceedings of the Computer Vision and Pattern Recognition ({CVPR})},
  year      = {2022},
  pages     = {3272--3282},
  doi       = {10.1109/CVPR52688.2022.00328}
}

@inproceedings{Bao2023USST,
  author    = {Bao, Wentao and Chen, Lele and Zeng, Libing and Li, Zhong and Xu, Yi and Yuan, Junsong and Kong, Yu},
  title     = {Uncertainty-aware State Space Transformer for Egocentric {3D} Hand Trajectory Forecasting},
  booktitle = {Proceedings of the {IEEE/CVF} International Conference on Computer Vision ({ICCV})},
  year      = {2023},
  pages     = {13656--13665},
  doi       = {10.1109/ICCV51070.2023.01260}
}

@inproceedings{Pasca2024TransFusion,
  author    = {Pasca, Razvan-George and Gavryushin, Alexey and Hamza, Muhammad and Kuo, Yen-Ling and Mo, Kaichun and Van Gool, Luc and Hilliges, Otmar and Wang, Xi},
  title     = {Summarize the Past to Predict the Future: Natural Language Descriptions of Context Boost Multimodal Object Interaction Anticipation},
  booktitle = {Proceedings of the {IEEE/CVF} Conference on Computer Vision and Pattern Recognition ({CVPR})},
  year      = {2024},
  pages     = {18286--18296},
  doi       = {10.1109/CVPR52733.2024.01731}
}

@inproceedings{Ma2025Diff-IP2D,
  author    = {Ma, Junyi and Chen, Xieyuanli and Xu, Jingyi and Wang, Hesheng},
  title     = {{Diff-IP2D}: Diffusion-Based Hand-Object Interaction Prediction on Egocentric Videos},
  booktitle = {Proceedings of the Intelligent Robots and Systems ({IROS})},
  year      = {2025},
  pages     = {4291--4298},
  doi       = {10.1109/IROS60139.2025.11247704}
}

@inproceedings{Soraki2026ObjectForesight,
  author    = {Soraki, Rustin and Bharadhwaj, Homanga and Farhadi, Ali and Mottaghi, Roozbeh},
  title     = {{ObjectForesight}: Predicting Future {3D} Object Trajectories from Human Videos},
  booktitle = {Proceedings of the European Conference on Computer Vision ({ECCV})},
  year      = {2026},
  volume    = {17062},
  pages     = {437--453},
  doi       = {10.1007/978-3-032-37098-3_24}
}

@article{Johansson2001EyeHandCoordination,
  author    = {Johansson, Roland S. and Westling, G{\"o}ran and B{\"a}ckstr{\"o}m, Anders and Flanagan, J. Randall},
  title     = {Eye--Hand Coordination in Object Manipulation},
  journal   = {The Journal of Neuroscience (J. Neurosci.)},
  year      = {2001},
  volume    = {21},
  number    = {17},
  pages     = {6917--6932},
  doi       = {10.1523/JNEUROSCI.21-17-06917.2001}
}

@inproceedings{Christen2024DiffH2O,
  author    = {Christen, Sammy and Hampali, Shreyas and Sener, Fadime and Remelli, Edoardo and Hodan, Tomas and Sauser, Eric and Ma, Shugao and Tekin, Bugra},
  title     = {{DiffH2O}: Diffusion-Based Synthesis of Hand-Object Interactions from Textual Descriptions},
  booktitle = {Proceedings of the {ACM} SIGGRAPH Asia Conference Papers ({SIGGRAPH Asia})},
  year      = {2024},
  articleno = {145},
  pages     = {1--11},
  doi       = {10.1145/3680528.3687563}
}

@inproceedings{He2025GazeGuided,
  author    = {He, Yufei and Zhang, Xucong and Stienen, Arno H. A.},
  title     = {Gaze-Guided {3D} Hand Motion Prediction for Detecting Intent in Egocentric Grasping Tasks},
  booktitle = {Proceedings of the {IEEE/RSJ} International Conference on Intelligent Robots and Systems ({IROS})},
  year      = {2025},
  pages     = {14580--14586},
  doi       = {10.1109/IROS60139.2025.11246573}
}

@inproceedings{Grauman2022Ego4D,
  author    = {Grauman, Kristen and Westbury, Andrew and Byrne, Eugene and Chavis, Zachary and Furnari, Antonino and Girdhar, Rohit and Hamburger, Jackson and Jiang, Hao and Liu, Miao and Liu, Xingyu and Martin, Miguel and Nagarajan, Tushar and Radosavovic, Ilija and Ramakrishnan, Santhosh Kumar and Ryan, Fiona and Sharma, Jayant and Wray, Michael and Xu, Mengmeng and Xu, Eric Zhongcong and Zhao, Chen and Bansal, Siddhant and Batra, Dhruv and Cartillier, Vincent and Crane, Sean and Do, Tien and Doulaty, Morrie and Erapalli, Akshay and Feichtenhofer, Christoph and Fragomeni, Adriano and Fu, Qichen and Gebreselasie, Abrham and Gonzalez, Cristina and Hillis, James and Huang, Xuhua and Huang, Yifei and Jia, Wenqi and Khoo, Weslie and Kolar, Jachym and Kottur, Satwik and Kumar, Anurag and Landini, Federico and Li, Chao and Li, Yanghao and Li, Zhenqiang and Mangalam, Karttikeya and Modhugu, Raghava and Munro, Jonathan and Murrell, Tullie and Nishiyasu, Takumi and Price, Will and Ruiz Puentes, Paola and Ramazanova, Merey and Sari, Leda and Somasundaram, Kiran and Southerland, Audrey and Sugano, Yusuke and Tao, Ruijie and Vo, Minh and Wang, Yuchen and Wu, Xindi and Yagi, Takuma and Zhao, Ziwei and Zhu, Yunyi and Arbelaez, Pablo and Crandall, David and Damen, Dima and Farinella, Giovanni Maria and Fuegen, Christian and Ghanem, Bernard and Ithapu, Vamsi Krishna and Jawahar, C. V. and Joo, Hanbyul and Kitani, Kris and Li, Haizhou and Newcombe, Richard and Oliva, Aude and Park, Hyun Soo and Rehg, James M. and Sato, Yoichi and Shi, Jianbo and Shou, Mike Zheng and Torralba, Antonio and Torresani, Lorenzo and Yan, Mingfei and Malik, Jitendra},
  title     = {{Ego4D}: Around the World in 3,000 Hours of Egocentric Video},
  booktitle = {Proceedings of the {IEEE/CVF} Conference on Computer Vision and Pattern Recognition ({CVPR})},
  year      = {2022},
  pages     = {18973--18990},
  doi       = {10.1109/CVPR52688.2022.01842}
}

@inproceedings{Banerjee2025HOT3D,
  author    = {Banerjee, Prithviraj and Shkodrani, Sindi and Moulon, Pierre and Hampali, Shreyas and Han, Shangchen and Zhang, Fan and Zhang, Linguang and Fountain, Jade and Miller, Edward and Basol, Selen and Newcombe, Richard and Wang, Robert and Engel, Jakob Julian and Hodan, Tomas},
  title     = {{HOT3D}: Hand and Object Tracking in {3D} from Egocentric Multi-View Videos},
  booktitle = {Proceedings of the {IEEE/CVF} Conference on Computer Vision and Pattern Recognition ({CVPR})},
  year      = {2025},
  pages     = {7061--7071},
  doi       = {10.1109/CVPR52734.2025.00662}
}

@article{Ma2026MADiff,
  author    = {Ma, Junyi and Chen, Xieyuanli and Bao, Wentao and Xu, Jingyi and Wang, Hesheng},
  title     = {{MADiff}: Motion-Aware {Mamba} Diffusion Models for Hand Trajectory Prediction on Egocentric Videos},
  journal   = {IEEE Transactions on Pattern Analysis and Machine Intelligence (TPAMI)},
  year      = {2026},
  volume    = {48},
  number    = {3},
  pages     = {3250--3267},
  doi       = {10.1109/TPAMI.2025.3633453}
}

@article{Ma2026Uni-Hand,
  author    = {Ma, Junyi and Bao, Wentao and Xu, Jingyi and Sun, Guanzhong and Zheng, Yu and Zhang, Erhang and Chen, Xieyuanli and Wang, Hesheng},
  title     = {{Uni-Hand}: Universal Hand Motion Forecasting in Egocentric Views},
  journal   = {IEEE Transactions on Pattern Analysis and Machine Intelligence (TPAMI)},
  year      = {2026},
  pages     = {1--18},
  doi       = {10.1109/TPAMI.2026.3692504}
}

@inproceedings{Liu2020FHOI,
  author    = {Liu, Miao and Tang, Siyu and Li, Yin and Rehg, James M.},
  title     = {Forecasting Human-Object Interaction: Joint Prediction of Motor Attention and Actions in First Person Video},
  booktitle = {Proceedings of the European Conference on Computer Vision ({ECCV})},
  year      = {2020},
  volume    = {12346},
  pages     = {704--721},
  doi       = {10.1007/978-3-030-58452-8_41}
}

@article{Dessalene2023EgoOMG,
  author    = {Dessalene, Eadom and Devaraj, Chinmaya and Maynord, Michael and Ferm{\"u}ller, Cornelia and Aloimonos, Yiannis},
  title     = {Forecasting Action Through Contact Representations From First Person Video},
  journal   = {IEEE Transactions on Pattern Analysis and Machine Intelligence (TPAMI)},
  year      = {2023},
  volume    = {45},
  number    = {6},
  pages     = {6703--6714},
  doi       = {10.1109/TPAMI.2021.3055233}
}

@article{Tong2026HandDiff,
  author    = {Tong, Jinguang and Zhang, Kaihao},
  title     = {{HandDiff}: Spatial--temporal diffusion model for hand pose forecasting},
  journal   = {Information Fusion (Inf. Fusion)},
  year      = {2026},
  volume    = {127},
  pages     = {103647},
  doi       = {10.1016/j.inffus.2025.103647}
}

@inproceedings{Cha2024Text2HOI,
  author    = {Cha, Junuk and Kim, Jihyeon and Yoon, Jae Shin and Baek, Seungryul},
  title     = {{Text2HOI}: Text-Guided {3D} Motion Generation for Hand-Object Interaction},
  booktitle = {Proceedings of the {IEEE/CVF} Conference on Computer Vision and Pattern Recognition ({CVPR})},
  year      = {2024},
  pages     = {1577--1585},
  doi       = {10.1109/CVPR52733.2024.00156}
}

@inproceedings{Zhang2025BimArt,
  author    = {Zhang, Wanyue and Dabral, Rishabh and Golyanik, Vladislav and Choutas, Vasileios and Alvarado, Eduardo and Beeler, Thabo and Habermann, Marc and Theobalt, Christian},
  title     = {{BimArt}: A Unified Approach for the Synthesis of {3D} Bimanual Interaction with Articulated Objects},
  booktitle = {Proceedings of the {IEEE/CVF} Conference on Computer Vision and Pattern Recognition ({CVPR})},
  year      = {2025},
  pages     = {27694--27705},
  doi       = {10.1109/CVPR52734.2025.02579}
}

@article{Tian2026GazeHOI,
  author    = {Tian, Jie and Ji, Ran and Yang, Lingxiao and Ni, Suting and Ma, Yuexin and Xu, Lan and Yu, Jingyi and Shi, Ye and Wang, Jingya},
  title     = {Gaze-guided Hand-Object Interaction Synthesis: Dataset and Method},
  journal   = {Transactions on Multimedia},
  year      = {2026},
  pages     = {1--11},
  doi       = {10.1109/TMM.2026.3710081}
}

@inproceedings{He2025SyncDiff,
  author    = {He, Wenkun and Liu, Yun and Liu, Ruitao and Yi, Li},
  title     = {{SyncDiff}: Synchronized Motion Diffusion for Multi-Body Human-Object Interaction Synthesis},
  booktitle = {Proceedings of the {IEEE/CVF} International Conference on Computer Vision ({ICCV})},
  year      = {2025},
  pages     = {11731--11743},
  doi       = {10.1109/ICCV51701.2025.01091}
}

@inproceedings{Prakash2025LatentAct,
  author    = {Prakash, Aditya and Lundell, Benjamin and Andreychuk, Dmitry and Forsyth, David and Gupta, Saurabh and Sawhney, Harpreet},
  title     = {How Do I Do That? Synthesizing {3D} Hand Motion and Contacts for Everyday Interactions},
  booktitle = {Proceedings of the {IEEE/CVF} Conference on Computer Vision and Pattern Recognition ({CVPR})},
  year      = {2025},
  pages     = {7026--7036},
  doi       = {10.1109/CVPR52734.2025.00659}
}

@inproceedings{Zhou2025MEgoHand,
  author    = {Zhou, Bohan and Zhan, Yi and Zhang, Zhongbin and Lu, Zongqing},
  title     = {{MEgoHand}: Multimodal Egocentric Hand-Object Interaction Motion Generation},
  booktitle = {Proceedings of the Conference on Neural Information Processing Systems ({NeurIPS})},
  year      = {2025},
  volume    = {38},
  pages     = {49464--49490},
  doi       = {10.52202/085713-1655}
}

@inproceedings{Hu2024GazeMotion,
  author    = {Hu, Zhiming and Schmitt, Syn and H{\"a}ufle, Daniel and Bulling, Andreas},
  title     = {{GazeMotion}: Gaze-guided Human Motion Forecasting},
  booktitle = {Proceedings of the {IEEE/RSJ} International Conference on Intelligent Robots and Systems ({IROS})},
  year      = {2024},
  pages     = {13017--13022},
  doi       = {10.1109/IROS58592.2024.10802548}
}

@inproceedings{Mondal2025GLAM,
  author    = {Mondal, Sounak and Sendhilnathan, Naveen and Zhang, Ting and Liu, Yue and Proulx, Michael and Iuzzolino, Michael Louis and Qin, Chuan and Jonker, Tanya R.},
  title     = {Gaze-Language Alignment for Zero-Shot Prediction of Visual Search Targets from Human Gaze Scanpaths},
  booktitle = {Proceedings of the {IEEE/CVF} International Conference on Computer Vision ({ICCV})},
  year      = {2025},
  pages     = {2738--2749},
  doi       = {10.1109/ICCV51701.2025.00263}
}

@inproceedings{Yan2024Voila-A,
  author    = {Yan, Kun and Wang, Zeyu and Ji, Lei and Wang, Yuntao and Duan, Nan and Ma, Shuai},
  title     = {{Voila-A}: Aligning Vision-Language Models with User's Gaze Attention},
  booktitle = {Proceedings of the Conference on Neural Information Processing Systems ({NeurIPS})},
  year      = {2024},
  volume    = {37},
  pages     = {1890--1918},
  doi       = {10.52202/079017-0060}
}

@inproceedings{Pani2025Gaze-VLM,
  author    = {Pani, Anupam and Yang, Yanchao},
  title     = {{Gaze-VLM}: Bridging Gaze and {VLMs} through Attention Regularization for Egocentric Understanding},
  booktitle = {Proceedings of the Conference on Neural Information Processing Systems ({NeurIPS})},
  year      = {2025},
  volume    = {38},
  pages     = {163544--163577},
  doi       = {10.52202/085713-5459}
}

@inproceedings{Peng2025EgoGazeVQA,
  author    = {Peng, Taiying and Hua, Jiacheng and Liu, Miao and Lu, Feng},
  title     = {In the Eye of {MLLM}: Benchmarking Egocentric Video Intent Understanding with Gaze-Guided Prompting},
  booktitle = {Proceedings of the Conference on Neural Information Processing Systems ({NeurIPS})},
  year      = {2025},
  volume    = {38},
  pages     = {121404--121424},
  doi       = {10.52202/085713-3659}
}

@inproceedings{Kwon2021H2O,
  author    = {Kwon, Taein and Tekin, Bugra and St{\"u}hmer, Jan and Bogo, Federica and Pollefeys, Marc},
  title     = {{H2O}: Two Hands Manipulating Objects for First Person Interaction Recognition},
  booktitle = {Proceedings of the {IEEE/CVF} International Conference on Computer Vision ({ICCV})},
  year      = {2021},
  pages     = {10118--10128},
  doi       = {10.1109/ICCV48922.2021.00998}
}

@inproceedings{Cheng2024EgoThink,
  author    = {Cheng, Sijie and Guo, Zhicheng and Wu, Jingwen and Fang, Kechen and Li, Peng and Liu, Huaping and Liu, Yang},
  title     = {{EgoThink}: Evaluating First-Person Perspective Thinking Capability of Vision-Language Models},
  booktitle = {Proceedings of the Computer Vision and Pattern Recognition ({CVPR})},
  year      = {2024},
  pages     = {14291--14302},
  doi       = {10.1109/CVPR52733.2024.01355}
}

@article{Romero2017MANO,
  author    = {Romero, Javier and Tzionas, Dimitrios and Black, Michael J.},
  title     = {Embodied Hands: Modeling and Capturing Hands and Bodies Together},
  journal   = {Transactions on Graphics (TOG)},
  year      = {2017},
  volume    = {36},
  number    = {6},
  pages     = {245:1--245:17},
  doi       = {10.1145/3130800.3130883}
}

@misc{Bai2025Qwen3VL,
  author    = {Bai, Shuai and Cai, Yuxuan and Chen, Ruizhe and Chen, Keqin and Chen, Xionghui and Cheng, Zesen and Deng, Lianghao and Ding, Wei and Gao, Chang and Ge, Chunjiang and Ge, Wenbin and Guo, Zhifang and Huang, Qidong and Huang, Jie and Huang, Fei and Hui, Binyuan and Jiang, Shutong and Li, Zhaohai and Li, Mingsheng and Li, Mei and Li, Kaixin and Lin, Zicheng and Lin, Junyang and Liu, Xuejing and Liu, Jiawei and Liu, Chenglong and Liu, Yang and Liu, Dayiheng and Liu, Shixuan and Lu, Dunjie and Luo, Ruilin and Lv, Chenxu and Men, Rui and Meng, Lingchen and Ren, Xuancheng and Ren, Xingzhang and Song, Sibo and Sun, Yuchong and Tang, Jun and Tu, Jianhong and Wan, Jianqiang and Wang, Peng and Wang, Pengfei and Wang, Qiuyue and Wang, Yuxuan and Xie, Tianbao and Xu, Yiheng and Xu, Haiyang and Xu, Jin and Yang, Zhibo and Yang, Mingkun and Yang, Jianxin and Yang, An and Yu, Bowen and Zhang, Fei and Zhang, Hang and Zhang, Xi and Zheng, Bo and Zhong, Humen and Zhou, Jingren and Zhou, Fan and Zhou, Jing and Zhu, Yuanzhi and Zhu, Ke},
  title     = {{Qwen3-VL} Technical Report},
  howpublished   = {arXiv preprint arXiv:2511.21631},
  year      = {2025},
  doi       = {10.48550/arXiv.2511.21631}
}

@inproceedings{Hu2022LoRA,
  author    = {Hu, Edward J. and Shen, Yelong and Wallis, Phillip and Allen-Zhu, Zeyuan and Li, Yuanzhi and Wang, Shean and Wang, Lu and Chen, Weizhu},
  title     = {{LoRA}: Low-Rank Adaptation of Large Language Models},
  booktitle = {Proceedings of the International Conference on Learning Representations ({ICLR})},
  year      = {2022}
}

@inproceedings{Salimans2022ProgressiveDistillation,
  author    = {Salimans, Tim and Ho, Jonathan},
  title     = {Progressive Distillation for Fast Sampling of Diffusion Models},
  booktitle = {Proceedings of the International Conference on Learning Representations ({ICLR})},
  year      = {2022}
}

@inproceedings{Peebles2023DiT,
  author    = {Peebles, William and Xie, Saining},
  title     = {Scalable Diffusion Models with Transformers},
  booktitle = {Proceedings of the {IEEE/CVF} International Conference on Computer Vision ({ICCV})},
  year      = {2023},
  pages     = {4172--4182},
  doi       = {10.1109/ICCV51070.2023.00387}
}

@inproceedings{Song2021DDIM,
  author    = {Song, Jiaming and Meng, Chenlin and Ermon, Stefano},
  title     = {Denoising Diffusion Implicit Models},
  booktitle = {Proceedings of the International Conference on Learning Representations ({ICLR})},
  year      = {2021}
}

@misc{Ho2022CFG,
  author    = {Ho, Jonathan and Salimans, Tim},
  title     = {Classifier-Free Diffusion Guidance},
  howpublished   = {arXiv preprint arXiv:2207.12598},
  year      = {2022},
  doi       = {10.48550/arXiv.2207.12598}
}

@misc{Hatano2025EgoH4,
  author  = {Hatano, Masashi and Zhu, Zhifan and Saito, Hideo and Damen, Dima},
  title   = {The Invisible {EgoHand}: {3D} Hand Forecasting through {EgoBody} Pose Estimation},
  howpublished = {arXiv preprint arXiv:2504.08654},
  year    = {2025}
}

@inproceedings{Tevet2023MDM,
  author    = {Tevet, Guy and Raab, Sigal and Gordon, Brian and Shafir, Yoni and Cohen-Or, Daniel and Bermano, Amit Haim},
  title     = {Human Motion Diffusion Model},
  booktitle = {International Conference on Learning Representations ({ICLR})},
  year      = {2023}
}

@inproceedings{Corona2020ContextAware,
  author    = {Corona, Enric and Pumarola, Albert and Aleny{\`a}, Guillem and Moreno-Noguer, Francesc},
  title     = {Context-Aware Human Motion Prediction},
  booktitle = {Proceedings of the {IEEE/CVF} Conference on Computer Vision and Pattern Recognition ({CVPR})},
  year      = {2020},
  pages     = {6992--7001}
}

@inproceedings{Cho2014GRU,
  author    = {Cho, Kyunghyun and van Merri{\"e}nboer, Bart and Gulcehre, Caglar and Bahdanau, Dzmitry and Bougares, Fethi and Schwenk, Holger and Bengio, Yoshua},
  title     = {Learning Phrase Representations using {RNN} Encoder--Decoder for Statistical Machine Translation},
  booktitle = {Proceedings of the Conference on Empirical Methods in Natural Language Processing ({EMNLP})},
  year      = {2014},
  pages     = {1724--1734},
  doi       = {10.3115/v1/D14-1179},
}

@inproceedings{Vaswani2017Transformer,
  author    = {Vaswani, Ashish and Shazeer, Noam and Parmar, Niki and Uszkoreit, Jakob and Jones, Llion and Gomez, Aidan N. and Kaiser, {\L}ukasz and Polosukhin, Illia},
  title     = {Attention Is All You Need},
  booktitle = {Advances in Neural Information Processing Systems ({NIPS})},
  year      = {2017},
  volume    = {30},
  pages     = {5998--6008}
}

@inproceedings{Furnari2019RULSTM,
  author    = {Furnari, Antonino and Farinella, Giovanni Maria},
  title     = {What Would You Expect? Anticipating Egocentric Actions With Rolling-Unrolling {LSTMs} and Modality Attention},
  booktitle = {Proceedings of the {IEEE/CVF} International Conference on Computer Vision ({ICCV})},
  year      = {2019},
  pages     = {6252--6261}
}

@inproceedings{Hu2026A3ffordHOI,
  author    = {Hu, Xurui and Qi, Xingqun and Yang, Bingkun and Su, Chen and Ru, Yiwei and Yin, Junhui and Sun, Muyi and Zhang, Man},
  title     = {{A$^{3}$fford-HOI}: Anatomy-Aligned Affordance Disentanglement for Fine-grained and Generalizable Hand Object Interaction},
  booktitle = {Proceedings of the International Joint Conference on Artificial Intelligence ({IJCAI})},
  year      = {2026},
  pages     = {1197--1205},
  doi       = {10.24963/ijcai.2026/134},
}

\end{document}